\documentclass{article}

\PassOptionsToPackage{numbers}{natbib}

\usepackage[final]{neurips_2024}

\usepackage[utf8]{inputenc} 
\usepackage[T1]{fontenc}    
\usepackage[breaklinks=true,colorlinks,bookmarks=false]{hyperref}
\usepackage{url}            
\usepackage{booktabs}       
\usepackage{amsfonts}       
\usepackage{nicefrac}       
\usepackage{microtype}      
\usepackage{amsthm}
\usepackage{amsmath}
\usepackage{graphicx}
\usepackage{amssymb}
\usepackage[table]{xcolor}
\usepackage{diagbox}
\usepackage{multirow}
\usepackage{pifont} 
\usepackage{bm}
\usepackage{subcaption}
\usepackage{color,xcolor}
\usepackage{caption}
\usepackage{wrapfig}
\usepackage[table]{xcolor}
\usepackage{colortbl}
\usepackage{enumitem}
\usepackage{makecell}
\usepackage{arydshln}
\usepackage{amsmath}
\usepackage{natbib}
\usepackage{wrapfig}
\usepackage{multirow}
\usepackage{algorithm, algpseudocode}
\usepackage{tcolorbox}
\usepackage{multicol}
\usepackage{algorithm}
\usepackage{tabularx}
\usepackage{algpseudocode}
\usepackage{threeparttable}
\usepackage{mathrsfs}

\hypersetup{
    colorlinks=true,
    linkcolor=red,
    citecolor=cyan,
    filecolor=magenta,      
    urlcolor=magenta,
    }

\title{Causal Deciphering and Inpainting in Spatio-Temporal Dynamics via Diffusion Model}

\author{
  Yifan Duan$^{1}$\thanks{Equal contribution}, Jian Zhao$^{2*}$\thanks{Corresponding authors}, pengcheng$^{5}$, Junyuan Mao$^{1*}$, Hao Wu$^{1}$, Jingyu Xu$^{3}$, \\
  \textbf{Shilong Wang$^{1}$, Caoyuan Ma$^{3}$, Kai Wang$^{4}$, Kun Wang$^{6\dagger}$, Xuelong Li}$^{2\dagger}$ \\
  $^{1}$University of Science and Technology of China, $^{2}$TeleAI, China Telecom, $^{3}$Wuhan University, \\
  $^{4}$National University of Singapore, $^{5}$Beijing Forestry University, $^{6}$Nanyang Technological University\\
  \texttt{\{duanyifan28,wslong1259,maojunyuan,wuhao2022\}@mail.ustc.edu.cn,} \\
  \texttt{\{kevinxu,macaoyuan\}@whu.edu.cn,pengcheng2022@bjfu.edu.cn,li@nwpu.edu.cn}, \\
  \texttt{wk520529wjh@gmail.com,\{E0823044,zhaojian90\}@u.nus.edu}
}

\begin{document}

\maketitle

\begin{abstract}
\vspace{-0.2em}
Spatio-temporal (ST) prediction has garnered a \textit{De facto} attention in earth sciences, such as meteorological prediction, human mobility perception. However, the scarcity of data coupled with the high expenses involved in sensor deployment results in notable data imbalances. Furthermore, models that are excessively customized and devoid of causal connections further undermine the {\fontfamily{lmtt}\selectfont generalizability} and {\fontfamily{lmtt}\selectfont interpretability}. To this end, we establish a framework for ST predictions from a causal perspective, termed {\fontfamily{lmtt}\selectfont CaPaint}, which targets to identify causal regions in data and endow model with causal reasoning ability in a two-stage process. Going beyond this process, we build on the front door adjustment as the theoretical foundation to specifically address the sub-regions identified as non-causal in the upstream phase. By using a fine-tuned unconditional Diffusion Probabilistic Model (DDPM) as the generative prior, we \underline{in-fill} the masks defined as environmental parts, offering the possibility of reliable extrapolation for potential data distributions. CaPaint overcomes the high complexity dilemma of optimal ST causal discovery models by reducing the data generation complexity from exponential to quasi-linear levels. Extensive experiments conducted on five real-world ST benchmarks demonstrate that integrating the {\fontfamily{lmtt}\selectfont \textbf{CaPaint}} concept allows models to achieve improvements ranging from 3.7\%$\sim$77.3\%. Moreover, compared to traditional mainstream ST augmenters, CaPaint underscores the potential of diffusion models in ST data augmentation, offering a novel paradigm for this field. Our project is available at \href{https://github.com/cpboost/123}{CaPaint}.

\end{abstract}

\vspace{-1em}
\section{Introduction}
\vspace{-0.5em}

Deep learning methodologies have achieved groundbreaking success across a wide array of spatio-temporal (ST) dynamics systems \cite{jin2023large,wang2020deep,Luo2024TimeseriesSA}, which include meteorological forecasting \cite{bi2022pangu,pathak2022fourcastnet,schultz2021can,wu2024earthfarsser}, wildfire spread modeling \cite{tam2022spatial,gerard2024wildfirespreadts}, intelligent transportation \cite{kaffash2021big,jin2023spatio,wang2022sfl}, and human mobility systems \cite{jin2023spatio,wu2023spatio}, to name just a few. 
Traditional ST dynamics approaches, based on first-principles \cite{burkle2021deep,pryor2009multiphysics}, often come with high computational costs. In contrast, ST dynamic analysis methods based on deep learning are not directly reliant on the explicit expression of physical laws but are data-driven \cite{jin2023large,wang2020deep,bi2022pangu,jin2023spatio}, relying on training models with large-scale observable datasets \cite{wang2022predrnn,shi2015convolutional, wu2024earthfarsser}. 


In a parallel vein, numerous efforts aim to incorporate physical laws into deep networks \cite{krishnapriyan2021characterizing, chen2022physics, raissi2019physics, karniadakis2021physics, wang2023scientific}, termed {\fontfamily{lmtt}\selectfont \textbf{Physics-Informed Neural Networks (PINNs)}}, which blend deep learning principles with physics to address challenges in scientific computing, particularly in fluid dynamics. {\fontfamily{lmtt}\selectfont \textbf{PINNs}} augment traditional neural network models by including a term in the loss function that accounts for the physical laws governing fluid dynamics, such as the Navier-Stokes equations \cite{constantin1988navier}. This ensures that the network’s predictions are not only consistent with empirical data but also comply with the fundamental principles of fluid dynamics. However, the off-the-shelf PINNs often suffer from limited generalization capabilities, primarily due to their \textit{customized loss function} designs and the \textit{neglect of specific network parameter} contexts \cite{takamoto2022cape, fotiadis2023disentangled}.

To date, the data-driven deep models are still dominant in ST dynamical systems, where the numerical simulation methods and PINNs generally lag behind. The reason may stem from the rise of large models \cite{achiam2023gpt, touvron2023llama, jin2023large} and the high costs associated with collecting ST data from sensors \cite{wu2024dynst, liu2024largest}, which creates a significant conflict between the increasing size of \textbf{\textit{data-hungry}} models and the \textbf{\textit{uneven, insufficient}} data collection. To this end, in the ST domain, there is looming research aimed at enhancing the causality and interpretability of models.

Unfortunately, research into causality within the field of ST dynamics is lagging. Although some work has considered causal design, due to specific domain constraints and architectural design, it can only enhance the tailor-made capabilities of the model for specific tasks \cite{xia2024deciphering,liu2011discovering}. Moreover, causal discovery tools \cite{di2020dominant,ebert2014causal} applied to ST systems often confront the ``curse of dimensionality'' issue during dimension reduction, despite their effectiveness in elucidating causal relationships from statistical data \cite{tibau2022spatiotemporal,nowack2020causal}. Furthermore, {\fontfamily{lmtt}\selectfont \textbf{NuwaDynamics}} \cite{wangnuwadynamics} for the first time proposed decomposing causal and non-causal regions in ST sequences and enhancing the robustness and generalizability of downstream model training by generating more potential distribution ST sequences through mixup \cite{zhang2017mixup}. CauSTG \cite{zhou2023maintaining} and CaST \cite{xia2024deciphering} address the issue of ST distribution shifts by implicitly modeling the time series embeddings and employing intervention techniques to observe these shifts.

Though promising, CauSTG \cite{zhou2023maintaining} and CaST \cite{xia2024deciphering} focus on modeling graph-related data, they lack an understanding of high-dimensional observational data (Dimension $\rm{D}<256$). {\fontfamily{lmtt}\selectfont \textbf{NuwaDynamics}}, on the other hand, explores all environments through backdoor adjustments \cite{pearl2009causality}, generating a vast number of sequences, which lead to nearly ${\cal O}( {T \times {\cal N}_E^{{\cal M}\left( {\rm{*}} \right)}} )$ training complexity ($T$ represents history time step, ${\cal N}_E$ and $\cal{M}(*)$ are the number of the environmental patches and mixup, respectively).

\begin{wrapfigure}{r}{8.5cm} \vspace{-1.0em}
 \centering
 \includegraphics[width=0.60\textwidth]{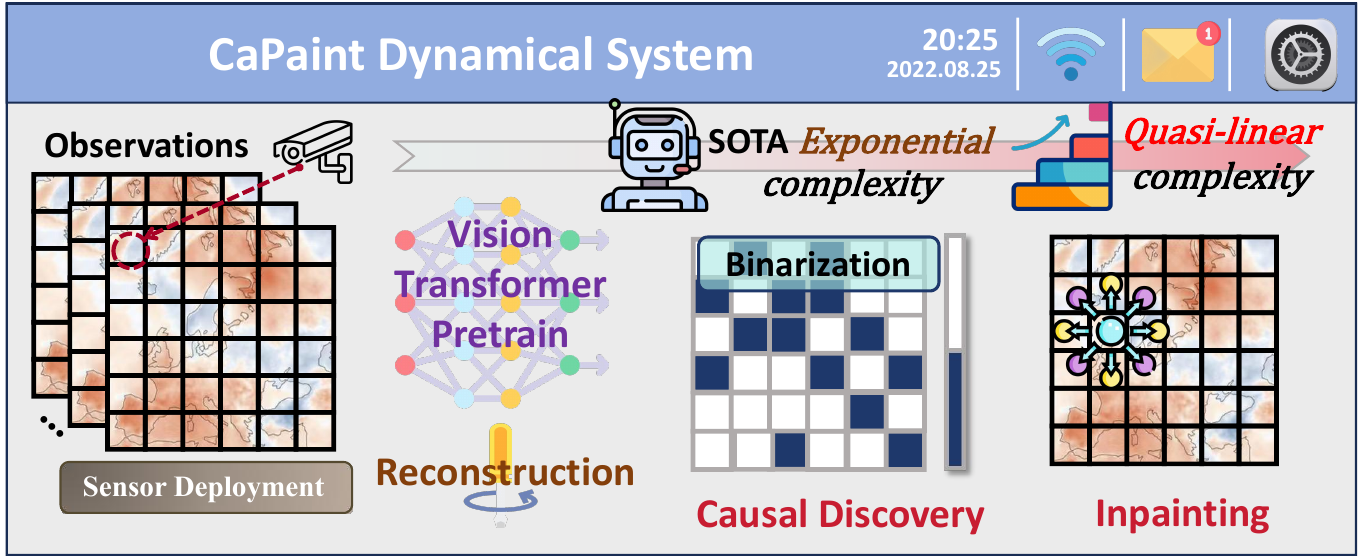}
  \vspace{-0.4em}
 \caption{Illustration of the CaPaint overview and advantage across SOTA ST causal model on complexity.}\label{intro:case}
   \vspace{-0.8em}
\end{wrapfigure}

In light of this, we propose a general causal structure plugin, termed \textit{CaPaint}, designed to decipher causal regions in ST data without adding extra computational cost, while intervening in non-causal areas to boost the model's generalizability and interpretability. Specifically, our method employs a straightforward approach to causal deciphering, utilizing a vision transformer architecture \cite{khan2022transformers} for self-supervised ST data reconstruction. During reconstruction, we leverage \textit{attention scores} from the self-attention mechanism \cite{han2022survey} to map onto important causal patches, thus endowing the model with interpretability. By ranking the entire set of importance scores, we define those with lower scores as environmental patches, which contribute minimally to the model. Building on this, we perform \textbf{causal interventions} in these environmental areas to aid the model in understanding more latent, complex, and imperceptible distributions, thereby enhancing the overall generalizability of the model (see Figure \ref{intro:case}). Concretely, we mask trivial regions and perform generation using DDPM \cite{ho2020denoising, karras2019style} fine-tuned on specific ST data, which can also be interpreted as a ST data inpainting approach.

\textbf{Insight.} \ding{182} CaPaint obeys the causal deciphering, and guided by the principle of frontdoor adjustment \cite{pearl2009causality,pearl2018book} from causal theory, CaPaint performs diffusion inpainting interventions on the environmental (non-causal) diffusion patches while reducing the temporal complexity to a manageable $\cal{O}(T \times {\cal N}_E)$ (from ${\cal O}( {T \times {\cal N}_E^{{\cal M}\left( {\rm{*}} \right)}} )$ in \cite{wangnuwadynamics}). \ding{183} CaPaint performs regional inpainting in a more natural manner, avoiding the predicament of repeatedly selecting and perturbing environmental patches. Through diffusion inpainting \cite{lugmayr2022repaint}, it generates images that are more aligned with the global distribution. \ding{184} CaPaint can be understood as a ST augmenter, offering a more rational concept of ST enhancement without disrupting the inherent distribution characteristics of space and time \cite{wang2020generalizing}. Our major contributions can be summarized as follow:

\begin{itemize}[leftmargin=*]
   \item In this paper, we introduce a novel causal structure plugin, CaPaint, which leverages the concept of frontdoor adjustment from causal theory. CaPaint enables various backbone models to learn from a broader distribution of data while providing enhanced interpretability for the models' predictions. 
   \vspace{-0.1em}
   
   \item By integrating diffusion generative models with ST dynamics, CaPaint selectively perturbs non-causal regions while maintaining the integrity of core causal areas. This approach generates valuable and reliable data for scenarios where high-quality data are scarce.
   \vspace{-0.1em}
   \item We conduct extensive experiments across five diverse and representative datasets from different domains, utilizing seven backbone models to assess the effectiveness of the CaPaint method. The empirical results demonstrate that CaPaint consistently enhances performance on all tested datasets and across all backbone models (3.7\%$\sim$77.3\%).

\end{itemize}

\vspace{-0.5em}
\section{Related work \& Technical Background} \label{related work}
\vspace{-0.3em}

\textbf{Spatio-temporal Predictive Learning:} Various architectures have achieved significant predictive performance in ST domain, which can primarily be categorized as follows: CNN-based models utilize convolutional layers to effectively capture spatial features \cite{mathieu2015deep,oh2015action,tulyakov2018mocogan,cheng2023rethinking}. RNN-based models, are capable of processing temporal sequence data and are well-suited for understanding temporal changes, showing excellent performance in the prediction of action continuity \cite{srivastava2015unsupervised,villegas2018hierarchical,wang2022predrnn,tan2023temporal}.
GNN-based models effectively capture spatial dependencies and temporal dynamics in data, making them suitable for complex tasks involving geographic locations and temporal changes \cite{Luo2024TimeseriesSA,Jiang2023UncertaintyQV,Li2022MiningSR,Gao2023UncertaintyAwarePG,zhang2024two,zhang2024graph,wang2024heterophilic, duan2024cat}. Transformer-based models employ self-attention mechanisms to process sequential data in parallel, enhancing the learning of long-term dependencies, and have been used for ST data prediction in complex scenarios \cite{bai2022rainformer,gao2022earthformer,wen2022transformers,wu2023pastnet,cheng2024nuwats,wang2023sst}. 


\textbf{Causal inference:} causal discovery algorithms, originally devised for unstructured random vectors \cite{shimizu2006linear,zheng2018dags}, have progressively been adapted for ST data analysis \cite{tibau2022spatiotemporal,nowack2020causal}. Within the extensive field of deep learning research, the study of causal inference aims to ensure a more stable and robust learning and reasoning paradigm. Recently, an array of techniques has been developed to delve into the nuances of causal features \cite{selvaraju2016grad,selvaraju2017grad,luo2020parameterized,ying2019gnnexplainer}, identifying and eliminating spurious correlations \cite{gulrajani2020search,koh2021wilds,sagawa2019distributionally}. 


\textbf{Generative models} especially diffusion-based model has gained significant popularity particularly in image and video generation \cite{ho2020denoising,shen2024boosting,shen2023advancing}. Sampling optimization algorithms have been used to accelerate the sampling process of diffusion models, significantly reducing the number of steps while improving efficiency. \cite{song2020denoising, lu2022dpm}. Additionally, generative models have also been applied to 3D scene generation and point cloud processing, as demonstrated in \cite{long2024wonder3d,karnewar2023holodiffusion,tang2024any2point,tang2024point,guo2023point,shen2024imagdressing}

\textbf{Image Inpainting} is a technique used to fill in missing or damaged parts of an image. This field can be broadly categorized into the following types. VAE-based methods: These methods leverage Variational Autoencoders to balance diversity and reconstruction  \cite{zhao2020uctgan,zheng2019pluralistic,Jiang2024IncompleteGL}. GAN-based methods: Since the introduction of Generative Adversarial Networks, these methods have been widely used for image inpainting \cite{richardson2021encoding,zhao2021large,pathak2016context}. Diffusion model-based methods: Diffusion models have recently shown outstanding performance in image inpainting 
 \cite{nichol2021glide,song2020score,saharia2022palette}.

\vspace{-0.5em}
\section{Methodology}
\vspace{-0.6em}
In this section, we systematically introduce causal structure plugin, \texttt{CaPaint}. Initially, we elucidate the methods employed in the upstream phase to delineate causal and non-causal regions (Sec \ref{sec:causal-deciphering}). Subsequently, we showcase the theoretical underpinnings supporting the \texttt{CaPaint} (Sec \ref{sec:SCM}). Building on this causal theory, we further engage in causal intervention within observational data (Sec \ref{sec:intervene}). Lastly, we demonstrate how sampling-enhanced ST observations can benefit the complexity of the model's \textit{on-device deployment} (Sec \ref{sec:Samling}).

\noindent\fbox{%
    \parbox{\linewidth}{%
        \textbf{Problem Formulation.} In ST settings, We represent ST observations as a sequence $\{X_t\}_{t=1}^T$, where each observation $X_t\in \mathbb{R}^{H \times W \times C_{\text{in}}}$ originates from these sequences. Our objective is to predict the trajectory for the forthcoming $K$ steps, denoted as $\{X_{t+1}\}_{t=T}^{T+K}$, with each future state $X_{t+k}$ mapped within $\mathbb{R}^{H \times W \times C_{\text{out}}}$. Here, $H$ and $W$ indicate the spatial grid dimensions, while $C_{\text{in}}$ and $C_{\text{out}}$ define the input and output dimensionality of the observations, respectively.
    }%
}
\vspace{-0.3em}

\begin{figure*}[!t]
  \centering
  \includegraphics[width=1.0\linewidth]{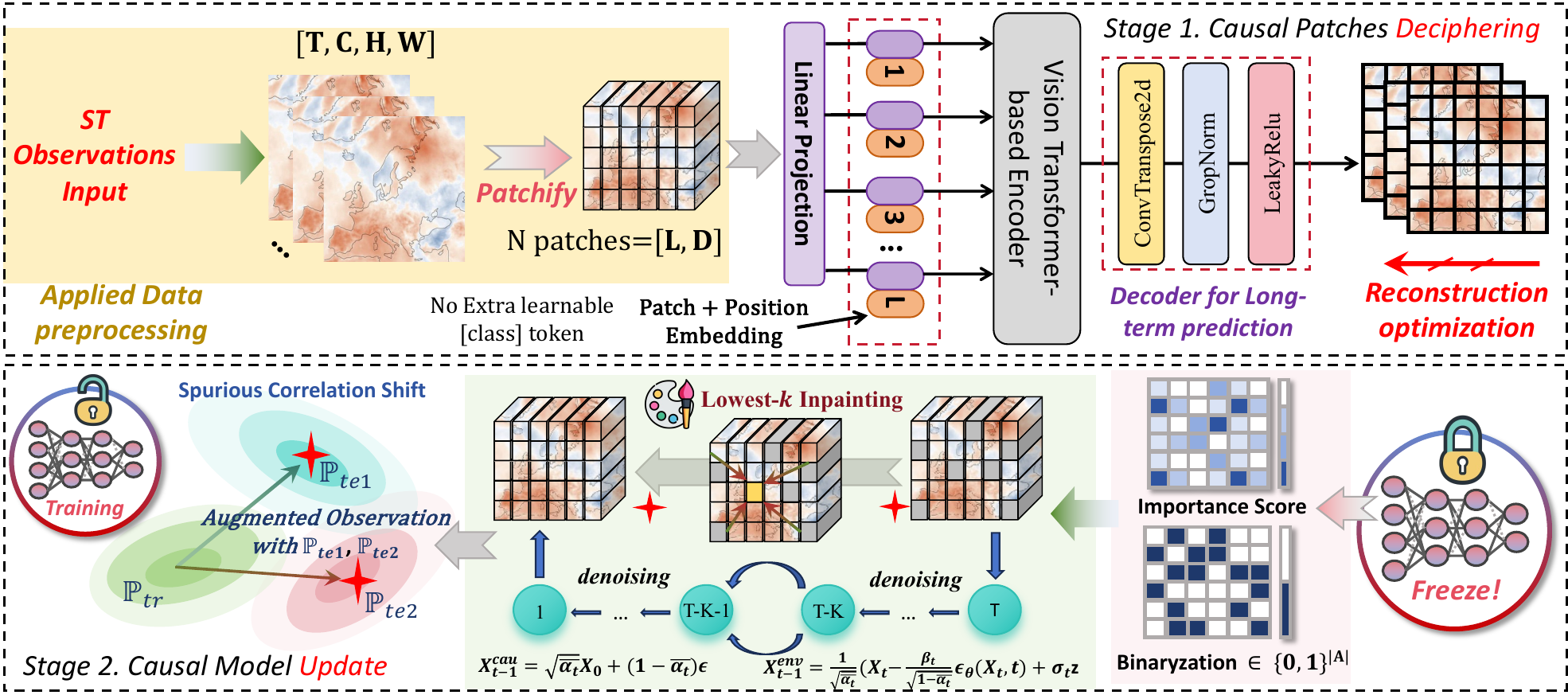}
  \vspace{-1.2em}
  \caption{The details of {\fontfamily{lmtt}\selectfont \textbf{CaPaint.}} (\textbf{\textit{Upper.}}) The initial phase of discovering causal patches. (\textbf{\textit{Bottom.}}) The update phase designed to eliminate spurious correlation shifts. Following the upstream training of the ViT, a diffusion model is trained in parallel. Using the identified causal patches as conditions, this generative model then performs inpainting for generating multiple sequences.
} 
  \label{fig:cap}
  \vspace{-1.0em}
\end{figure*}

\vspace{-0.4em}
\subsection{Causal Deciphering}
\label{sec:causal-deciphering}
\vspace{-0.4em}

To find the causal (non-causal) patches with \textbf{no labels}, we employ a self-supervised \texttt{reconstruction} approach based on the Vision Transformer (ViT) \cite{dosovitskiy2020image} to identify key regions within ST observations. ViT segments the image into multiple patches and calculates the relationships between them using a self-attention mechanism. Due to no label property, we intentionally omit the use of the  {\fontfamily{lmtt}\selectfont \textbf{[Cls]}} token in classification task and send data into ViT for encouraging \textit{``local-to-global''} reconstruction.

Specifically, each ST data $X_t$, is divided into $N = HW/p^2$ patches, where each patch $x_t^{patch} \in \mathbb{R}^{N \times (p^2 \times C_{\text{in}})}$, with $(H,W)$ being the resolution of the original ST data and $(p,p)$ the resolution of each patch. Subsequently, each patch is mapped to a $\rm{D}$-dimensional token through a learnable linear layer, incorporating position embedding to enhance the model's sensitivity to positional information. These tokens are then fed into successive $L$ stacked transformer blocks, as described in Equation~\ref{eq:bit}:  
\begin{equation} \small
\label{eq:bit}
    L \times \left( {\rm{X'}} = \underbrace{ {\rm{X}} + \rm{MSA}\left( LN\left( X \right) \right)}_{\textrm{Multi-head Attention}} \;\; \Rightarrow \;\;
    X_{\rm{out}} = {\rm{X'}} + \underbrace{\rm{MLP}\left( LN\left( X' \right) \right)}_{\textrm{Residual Connection}} \right)
\end{equation}
where LN denotes layer normalization, and MLP represents multi-layer perceptron. The upstream self-supervised reconstruction task enables the model to learn intrinsic property of ST data. Navigating the MSA mechanism \cite{vaswani2017attention,Yang2024FaiMAFI}, each patch $x_t^{patch}$ derived from the ST observation \(X_t\) is transformed into queries \(q\), keys \(k\), and values \(v\), and then calculates the relevance of each patch to others, forming a weighted representation that focuses on the most informative parts. The attention weights \(A_{i,j}^h\) stored in the attention map $A^h$ in each head are computed using the scaled dot-product:
\vspace{-0.4em}
\begin{equation} \small
{\rm{set}}\left\{ {Q,K,V} \right\} = {X_t}{\psi _{tr}},\quad {A^h} = {\rm{Softmax}}\left( {\frac{{Q{K^T}}}{{\sqrt {{D_h}} }}} \right) = {\left( {\begin{array}{*{20}{c}}
{A_{1,1}^h}& \cdots &{A_{1,N}^h}\\
 \vdots & \ddots & \vdots \\
{A_{N,1}^h}& \cdots &{A_{N,N}^h}
\end{array}} \right)_{A_{\left\{ {i,j} \right\} \in 1 \to N}^h}}
\end{equation}
\vspace{-0.7em}


where ${\psi _{tr}} \in \mathbb{R}^{N \times 3D_h}$ are the parameter matrices, $D_h$ represents the dimension of each head, \(Q\), \(K\), and \(V\) collectively denote the sets of queries \(q\), keys \(k\), and values \(v\). In our approach, the determination of causal patches, is driven by an analysis of the attention maps $A$. Each row in an attention map is normalized and represents the importance of other patches relative to the current patch $x_t^{i}$. However, to ascertain the overall importance of each patch across the entire input, \textit{we aggregate the contributions by summing the values along the \underline{columns} of the $A$}.  To integrate insights across multiple heads, we sum these measures across all heads and then normalize the resultant vector to derive a comprehensive importance score for each patch:


\vspace{-0.5em}
\begin{equation} \small
S \in \mathbb{R}^N = \text{Softmax}\left(\sum_{h=1}^{H} \sum_{i=1}^{N} A_{i,j}^{h}\right)
\end{equation}
\vspace{-0.4em}

where $S$ represents the normalized importance score vector, $A_{i,j}^h \in A$ denotes the attention that $x_t^{i}$ pays to $x_t^{j}$ for each head, $H$ is the number of heads. We sort the importance scores into $S$ and select the patches corresponding to the lowest $K$ scores as environmental patches that are stored in $O_e$. The remaining patches are considered causal patches $O_c$:
\vspace{-0.5em}
\begin{equation} 
O_{c} = {\rm{Topk}}\left( \lceil\mathcal{C}(S) \times \epsilon\% \rceil, \;\;{\underset{S_{i} \in S}{\arg\max}{\left\{ {\rm{set}}\left( \Psi\left( X_{t} \right) \right) \right)\}}} \right)
\end{equation}
\vspace{-0.5em}

where \(C(S)\) is the counting function, \(\epsilon\) represents the proportion of patches selected as causal, and \(\Psi(X_t)\) denotes the set of patches in the ST observation \(X_t\). We identify the causal patches by locating the indices with the highest values in \(S\) and define the non-causal parts as the environmental parts. Our goal is to perform causal interventions on the environmental parts.


\vspace{-0.4em}
\subsection{Backdoor Adjustment v.s Frontdoor Adjustment}
\label{sec:SCM}
\vspace{-0.4em}

\begin{wrapfigure}{r}{6.5cm} \vspace{-1.0em}
 \centering
 \includegraphics[width=0.45\textwidth]{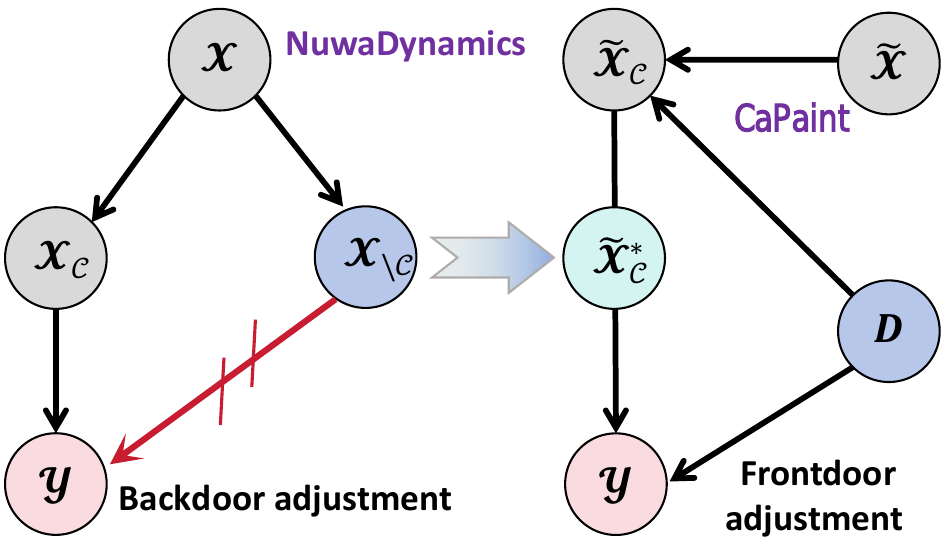}
  \vspace{-0.4em}
 \caption{Different SCM architectures of {\fontfamily{lmtt}\selectfont \textbf{SOTA}} and CaPaint.}\label{method:SCM}
   \vspace{-0.8em}
\end{wrapfigure}

To address issues of ST data scarcity and poor transferability, we examine the evaluation process using a Structural Causal Model (SCM) \cite{pearl2018book}, as shown in Fig \ref{method:SCM}. We represent abstract data variables by nodes, with directed links symbolizing causality. The SCM illustrates the interaction among variables through a graphical definition of causation, demonstrating the interconnected nature of these elements. As depicted in the left part, {\fontfamily{lmtt}\selectfont \textbf{NuwaDynamics}} employs the backdoor adjustment to enhance the generalization performance of the model:

\vspace{-0.6em}
\begin{itemize}[leftmargin=*]
    \item[\ding{224}] ${{\mathcal X}_{\mathcal C}} \leftarrow {\mathcal X} \to {{\mathcal X}_{\backslash {\mathcal C}}}$. The input ${\mathcal X}$ consists of two disjoint parts ${{\mathcal X}_{\mathcal C}}$ (causal part) and ${{\mathcal X}_{\backslash {\mathcal C}}}$ (environmental or trivial part).

     \item[\ding{224}] ${{\mathcal X}_{\mathcal C}} \to {\mathcal Y} \nleftarrow {{\mathcal X}_{\backslash {\mathcal C}}}$. Here, ${{\mathcal X}_{\mathcal C}}$ represents the sole endogenous parent that determines the ground truth $\mathcal Y$. However, in practical scenarios, ${{\mathcal X}_{\backslash {\mathcal C}}}$ is also employed in predicting $\mathcal Y$, which leads to the formation of spurious associations.
     
\end{itemize}
\vspace{-0.6em}

\noindent In general, a model ${{\mathcal F}_\emptyset }$ trained using Empirical Risk Minimization (ERM) often struggles to generalize to the test data $\mathcal{D}_{te} \sim \mathbb{P}_{te}$. Such distribution shifts are often induced by variations in environmental patches. Hence, addressing the confounding effect caused by the environmental confounder is crucial. Backdoor adjustment techniques are employed to perturb the environmental components, thereby enhancing the model’s potential to observe a broader range of latent distributions by forcibly perturbing the environmental variables ${{\mathcal X}_{\backslash {\mathcal C}}}$ (referred to as the \textbf{do-calculus} \cite{pearl2009causality} operator). Unfortunately, \textcolor{black}{\ding{182}} traversing all environmental variables is quite challenging. Although {\fontfamily{lmtt}\selectfont \textbf{NuwaDynamics}} uses Gaussian sampling to mitigate the issue of complexity, controlling Gaussian sampling in temporal sequence operations is particularly difficult. It requires meticulous adjustment of mean and variance to ensure a balance between the number of environmental samples and the training burden. \textcolor{black}{\ding{183}} Worse still, by traversing all environments, it likely violates underlying properties, including distribution shift content and nonexistent scenarios \cite{wu2022deconfounding}. To address this issue, we employ front-door adjustment, as illustrated in the right half of the Fig \ref{method:SCM}:
\vspace{-0.5em}
\begin{itemize}[leftmargin=*]
    \item ${\tilde {\mathcal X}_{\mathcal C}} \leftarrow {\mathcal D} \to {\mathcal Y}$. In this structure, ${\mathcal D}$ serves as a confounder, creating a misleading path between ${\tilde {\mathcal X}_{\mathcal C}}$ and ${\mathcal Y}$. Here, ${\tilde {\mathcal X}_{\mathcal C}}$ represents the causal component within $\tilde {\mathcal X}$.

     \item ${\tilde {\mathcal X}_{\mathcal C}} \to \tilde {\mathcal X}_{\mathcal C}^* \to {\mathcal Y}$. $\tilde {\mathcal X}_{\mathcal C}^*$ acts as the surrogate variable of ${\tilde {\mathcal X}_{\mathcal C}}$ and completes ${\tilde {\mathcal X}_{\mathcal C}}$ to align it with the data distribution. Initially, it derives from and encompasses ${\tilde {\mathcal X}_{\mathcal C}}$. Specifically, it envisions the potential complete observations that should exist when observing the sub-counterpart ${\tilde {\mathcal X}_{\mathcal C}}$. Additionally, $\tilde {\mathcal X}_{\mathcal C}^*$ adheres to the data distribution and upholds the intrinsic knowledge of graph properties, thus eliminating any link between ${\mathcal D}$ and $\tilde {\mathcal X}_{\mathcal C}^*$. Consequently, $\tilde {\mathcal X}_{\mathcal C}^*$ is well-suited to act as the mediator, which in turn influences the model's predictions ($\to {\mathcal Y}$).
\end{itemize}
\vspace{-0.5em}

 In our front-door adjustment framework, we utilize \textbf{do-calculus} on the variable ${\tilde {\mathcal X}_{\mathcal C}}$ to eliminate the spurious correlations introduced by ${\mathcal D} \to {\mathcal Y}$. Specifically, we achieve this by summing over potential surrogate observations $\tilde {X}_{\mathcal C}^*$. This approach allows us to connect two identifiable partial effects: ${\tilde {\mathcal X}_{\mathcal C}} \to \tilde {\mathcal X}_{\mathcal C}^*$ and $\tilde {\mathcal X}_{\mathcal C}^* \to {\mathcal Y}$:
 \vspace{-0.5em}
\begin{equation} \footnotesize
 \begin{aligned}
    & P\left( {{\mathcal Y}{\rm{|}}do\left( {{{\tilde {\mathcal X}}_{\mathcal C}} = {{\tilde X}_{\mathcal C}}} \right)} \right)  = \mathop \sum \limits_{\tilde X_{\mathcal C}^*} P\left( {{\mathcal Y}{\rm{|}}do\left( {\tilde {\mathcal X}_{\mathcal C}^* = \tilde X_{\mathcal C}^*} \right)} \right)P\left( {\tilde {\mathcal X}_{\mathcal C}^* = \tilde X_{\mathcal C}^*{\rm{|}}do\left( {{{\tilde {\mathcal X}}_{\mathcal C}} = {{\tilde X}_{\mathcal C}}} \right)} \right) \\ 
    & = {\sum\limits_{{\overset{\sim}{X}}_{\mathcal{C}}^{*}}{\sum\limits_{{\overset{\sim}{X}}_{\mathcal{C}}^{'}}{P\left( \mathcal{Y} \middle| {{\overset{\sim}{\mathcal{X}}}_{\mathcal{C}}^{*} = {\overset{\sim}{X}}_{\mathcal{C}}^{*};{\overset{\sim}{\mathcal{X}}}_{\mathcal{C}} = {\overset{\sim}{X}}_{\mathcal{C}}^{'}} \right)}}}P\left( {\overset{\sim}{\mathcal{X}}}_{\mathcal{C}} = {\overset{\sim}{X}}_{\mathcal{C}}^{'} \right)P\left( {{\overset{\sim}{\mathcal{X}}}_{\mathcal{C}}^{*} = {\overset{\sim}{X}}_{\mathcal{C}}^{*}} \middle| {do\left( {{\overset{\sim}{\mathcal{X}}}_{\mathcal{C}} = {\overset{\sim}{X}}_{\mathcal{C}}} \right)} \right) \\
    & = {\sum\limits_{{\overset{\sim}{X}}_{\mathcal{C}}^{*}}{\sum\limits_{{\overset{\sim}{X}}_{\mathcal{C}}^{'}}{P\left( \mathcal{Y} \middle| {{\overset{\sim}{\mathcal{X}}}_{\mathcal{C}}^{*} = {\overset{\sim}{X}}_{\mathcal{C}}^{*};{\overset{\sim}{\mathcal{X}}}_{\mathcal{C}} = {\overset{\sim}{X}}_{\mathcal{C}}^{'}} \right)}}}P\left( {\overset{\sim}{\mathcal{X}}}_{\mathcal{C}} = {\overset{\sim}{X}}_{\mathcal{C}}^{'} \right)P\left( {{\overset{\sim}{\mathcal{X}}}_{\mathcal{C}}^{*} = {\overset{\sim}{X}}_{\mathcal{C}}^{*}} \middle| {{\overset{\sim}{\mathcal{X}}}_{\mathcal{C}} = {\overset{\sim}{X}}_{\mathcal{C}}} \right)
 \end{aligned}
\end{equation}
 \vspace{-0.5em}

\noindent $P\left( {\tilde {\mathcal X}_{\mathcal C}^*{\rm{|}}do\left( {{{\tilde {\mathcal X}}_{\mathcal C}} = {{\tilde X}_{\mathcal C}}} \right)} \right) = P\left( {\tilde {\mathcal X}_{\mathcal C}^*{\rm{|}}{{\tilde {\mathcal X}}_{\mathcal C}} = {{\tilde X}_{\mathcal C}}} \right)$ holds as ${\tilde {\mathcal X}_{\mathcal C}}$ is the only parent of $\tilde {\mathcal X}_{\mathcal C}^*$. With data pair $({\tilde {\mathcal X}_{\mathcal C}}, \tilde {\mathcal X}_{\mathcal C}^*)$, we can feeding the surrogate observations $\tilde {\mathcal X}_{\mathcal C}^*$ into our ST framework, conditional on the ${\tilde {\mathcal X}_{\mathcal C}}$, to estimate $P\left( \mathcal{Y} \middle| {{\overset{\sim}{\mathcal{X}}}_{\mathcal{C}}^{*} = {\overset{\sim}{X}}_{\mathcal{C}}^{*};{\overset{\sim}{\mathcal{X}}}_{\mathcal{C}} = {\overset{\sim}{X}}_{\mathcal{C}}^{'}} \right)$. Compared to previous work {\fontfamily{lmtt}\selectfont \textbf{NuwaDynamics}}, CaPaint utilizes causal regions to generate global surrogate variables in a more rational manner, circumventing the cumbersome need to traverse environmental variables inherent in backdoor adjustments. \textbf{In fact, backdoor adjustments often likely violate underlying properties, leading to the generation of non-existent data distributions.} The broader scenarios of CaPaint will be detailed in Appendix \ref{app:c}.

\vspace{-0.4em}
\subsection{Causal Intervention via Diffusion Inpainting} 
\label{sec:intervene}
\vspace{-0.4em}

Building on the principles of causal analysis outlined above, we proceed to perform interventions on the environmental patches using diffusion inpainting, which enables us to manipulate the environmental areas. Initially, given the unique complexities of ST datasets, we \textit{fine-tune} the diffusion parameters to adapt seamlessly to the domain-specific challenges, which enhances the accuracy of our interventions on environmental patches. Diffusion models learn the distribution of data through a forward noise addition process and a reverse denoising process:
\begin{equation} \small
q(X_t \mid X_{t-1}) = \mathcal{N}(X_t; \sqrt{1 - \beta_t}X_{t-1}, \beta_t I), \quad
p_{\theta}(X_{t-1} \mid x_t) = \mathcal{N}(X_{t-1}; \mu_{\theta}(X_t, t), \Sigma_{\theta}(X_t, t))
\end{equation}

where $X_t$ represents the data state at time step $t$, undergoing a transformation from its previous state $x_{t-1}$, $\beta_t$ controls the variance of the noise added at each step in the forward process, $\mu_{\theta}$ and $\Sigma_{\theta}$ are neural network outputs that approximate the mean and covariance, respectively. The fine-tuning objective of the diffusion process is designed to approximate the data distribution more accurately. Specifically, the training objective for diffusion models, denoted as $\epsilon_\theta$, which predicts the noise, is typically defined as a simplified version of the variational bound:
\begin{equation}
L_{\text{simple}}=\mathbb{E}_{X_{0},\boldsymbol{\epsilon}\sim \mathcal{N}(\mathbf{0}, \mathbf{I}), \boldsymbol{c}, t} \| \boldsymbol{\epsilon}- \boldsymbol{\epsilon}_\theta\big(X_t, \boldsymbol{c}, t\big)\|^2
\end{equation}
where $c$ is the condition information. In this paper, we perform inpainting on the environmental patches of ST data. Inspired by \cite{lugmayr2022repaint}, we generate a mask image for each  ST data where the causal patches are black and the environmental patches are white. By independently sampling the causal and environmental patches and applying the diffusion inpainting process, we are able to generate augmented ST observation data. The detailed algorithmic process is shown in Appendix \ref{app:a}. 
\vspace{-0.4em}
\begin{equation}
X_{t-1}^{cau} = \sqrt{\bar\alpha_t} X_{0} + (1 - \bar\alpha_t) \epsilon, \quad X_{t-1}^{env} = \frac{1}{\sqrt{\alpha_t}} \left(X_t - \frac{\beta_t}{\sqrt{1-\bar\alpha_t}} \epsilon_\theta(X_t, t) + \sigma_t z\right) 
\end{equation}
\vspace{-0.4em}
\begin{equation}
\label{equ:merge}
X_{t-1} = m \odot X_{t-1}^{cau} +  (1-m) \odot X_{t-1}^{env}
\end{equation}

where $X^{cau}$ and $X^{env}$ denote causal patches and environmental patches, $m$ is a binary mask matrix, $\alpha_t$ represents the scaling factor at each diffusion step, determining the variance retained in the transition from $X_{t-1}$ to $X_t$. The cumulative product $\bar{\alpha}_t = \prod_{i=1}^t \alpha_i$ represents the accumulated scaling effect from the $T=0$ to step $t$. Equation \ref{equ:merge} illustrates the merging of environmental patches and causal patches. Finally, the enhanced ST observation data are stored within our temporal sequence repository to bolster the downstream backbone.

\vspace{-0.4em}
\subsection{ST Sequence Sampling Modeling}
\label{sec:Samling}
\vspace{-0.4em}
Previous work \cite{wangnuwadynamics} assumed that the closer the time point is to the present, the greater its influence, and thus used Gaussian sampling to select more ST data closer to the current time point. However, we argue that uniform sampling can better enhance the model's generalization ability. To enhance computational efficiency while ensuring prediction accuracy, we employ a ST sequence modeling approach that samples at each time point with a fixed probability controlled by the hyperparameter \( p \). This method allows us to sample from both original and generated data at each time point, thereby creating a new spatiotemporal sequence. We use two hyperparameters: \( p \), which controls the sampling probability, and \( r \), which determines the number of generated spatiotemporal sequences, achieving an optimal balance between computational efficiency and prediction accuracy. The specific sampling process can be represented by the following equation:

\vspace{-0.5em}
\begin{equation} \small
\begin{aligned}
X'_{t} = \text{Sample}(X_t, p, r)
\end{aligned}
\end{equation}

where \( X_t \) represents the collection of original and generated data at time point \( t \), and \(\text{Sample}(X_t, p)\) denotes the dataset obtained by sampling from \( X_t \) with probability \( p \). The hyperparameter \( p \) is directly set as the sampling probability, while \( r \) is used to specify the number of generated ST sequences.


\vspace{-0.4em}
\section{Experiments}\label{sec:exp}
\vspace{-0.4em}
In this section, we will validate the effectiveness of our proposed causal structure plugin, CaPaint. We design four research questions (RQs) to comprehensively evaluate the performance of CaPaint: \textbf{RQ1:} Does CaPaint effectively enhance model performance and applicability? \textbf{RQ2:} How does CaPaint perform in data-scarce scenarios? \textbf{RQ3:} How does the performance of CaPaint compare with other augmentation methods? \textbf{RQ4:} Is CaPaint effective for long-term time step predictions? Through these research questions, we aim to validate the effectiveness and advantages of CaPaint in handling ST data from multiple perspectives.

\vspace{-0.4em}
\subsection{Experimental settings}
\label{sec:settings}
\vspace{-0.4em}
\textbf{Datasets.} 
We extensively evaluate our proposal using a diverse range of benchmark datasets spanning multiple fields, include FireSys \cite{chen2022automated}, SEVIR \cite{veillette2020sevir}, Diffusion reaction system (DRS) \cite{chen2021discovering}, KTH \cite{schuldt2004recognizing} and TaxiBJ+ \cite{liang2021fine}. Specifically, FireSys represents fire dynamics, SEVIR covers meteorological events, DRS involves physical control systems, KTH focuses on human motion dynamics, and TaxiBJ+ is a transportation dataset. Detailed information can be found in the Appendix \ref{app:b}.

\textbf{Backbones and Metrics}
To validate the generalizability of CaPaint, we select multiple model frameworks for our experiments, including the classic model like ConvLSTM \cite{shi2015convolutional}, PredRNN-V2 \cite{wang2022predrnn}, Vision Transformer (ViT) \cite{dosovitskiy2020image}, MAU \cite{chang2021mau}, the efficiency-focused SimVP \cite{gao2022simvp}, and some of the latest models like MmvP \cite{zhong2023mmvp} and Earthfarsser \cite{wu2024earthfarsser}. Our evaluation metrics include mean absolute error (MAE), mean squared error (MSE), and structural 
similarity index measure (SSIM). Detailed information can be found in the Appendix \ref{metrics}.

\vspace{-1mm}
\begin{figure}[htbp]
    \centering
    \includegraphics[width=\textwidth]{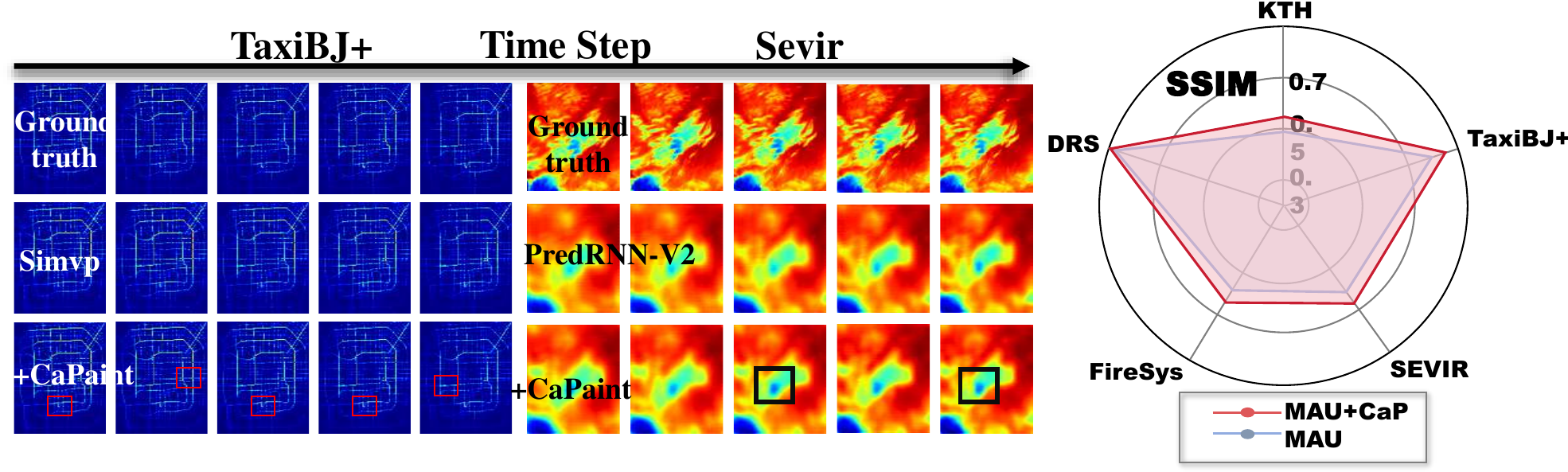}
    \caption{Visualization of prediction results for TaxiBJ+ and SEVIR datasets. The left side shows the predicted results of the last 5 frames for TaxiBJ+. The middle presents the results of long-term predictions for SEVIR, displaying the last five frames from step 10 $\rightarrow$ step 20. The right side compares SSIM metrics with and without the incorporation of CaPaint.}
    \label{fig:ob}
\end{figure}
\vspace{-0.2em}

\vspace{-0.4em}
\subsection{Evaluating the Efficacy of CaPaint (RQ1 \& RQ4)}
\vspace{-0.4em}

In this section, we conduct extensive experiments to demonstrate the effectiveness of the \textit{CaPaint} method. For Transformer architectures, we can directly transfer the model parameters trained in upstream tasks, thereby achieving efficient downstream training. For non-Transformer architectures, we focus on transferring the data itself to train the downstream models. The data presented in the Tab \ref{tab:main} show the performance improvements achieved by generating \textbf{only one} single generalized copy for each ST sequence. As shown in the Table \ref{tab:main}, we can list the \textbf{Obs}ervations: 

\noindent\textbf{Obs 1. +CaPaint consistently leads w/o Capaint settings across all datasets.}~As shown in Table~\ref{tab:main} and the right side of Fig \ref{fig:ob}, we can easily observe that introducing +CaPaint significantly improves model performance on MAE, MSE and SSIM metrics across all datasets. For example, with the ViT model on TaxiBJ+, MAE drops from 16.59 $\rightarrow$ 14.54, MSE from 11.40 $\rightarrow$ 8.89; On Diffusion Reaction Systems, MAE significantly decreases from 13.59 $\rightarrow$ 7.52, MSE from 6.21 $\rightarrow$ 1.41. This shows CaPaint's effectiveness in boosting performance in various domains.

\noindent\textbf{Obs 2. +CaPaint  enhances model local insights ST scenarios.}~By analyzing the left side of Figure~\ref{fig:ob}, we clearly see that the +CaPaint  effectively reduces the model's prediction loss. Moreover, it is observed that +CaPaint provides more accurate predictions in finer details, closely aligning with the actual result curves. This demonstrates CaPaint's capability to enhance prediction accuracy and reliability, ensuring that the forecasts closely mirror real-world outcomes.

\noindent\textbf{Obs 3. +CaPaint remains effective in long-Term ST predictions.}~By analyzing the middle of Figure~\ref{fig:ob}, we observe that +CaPaint continues to demonstrate its effectiveness in long-term time step predictions for ST tasks. For instance, the details in the SEVIR dataset predictions improve significantly, indicating that CaPaint is still applicable and beneficial in challenging ST tasks.

\begin{table}[t] \scriptsize
\caption{\footnotesize This table showcases the results (five runs) differences between using the CaPaint concept (+CaP) and not using it (Ori) across various datasets. All MAE and MSE values are multiplied by 100. \textcolor{blue}{Blue} and \textcolor{red}{Red} backgrounds indicate the percentage improvement (reduction) in MAE and MSE, respectively.} 
\label{tab:main}
\setlength{\tabcolsep}{0.80pt}
\begin{center}
\def \arraystretch{1.0}
 \begin{tabularx}{\textwidth}{cccXXXXXXXXc} 
 \toprule
    \multirow{2}{*}{\small \bf \makecell[c]{Backbone \\ ($10\rightarrow10$)}} & \multirow{2}{*}{\small \bf Metric} & \multicolumn{2}{c}{\scriptsize \bf TaxiBJ+ } & \multicolumn{2}{c}{\scriptsize \bf KTH}  &  \multicolumn{2}{c}{\scriptsize \bf SEVIR }  & 
    \multicolumn{2}{c}{\scriptsize \bf DRS   }  & 
    \multicolumn{2}{c}{\scriptsize \bf FireSys }   \\    
    \cmidrule(r){3-4} \cmidrule(r){5-6} \cmidrule(r){7-8} \cmidrule(r){9-10}  \cmidrule(r){11-12} 
     && \scriptsize \bf Ori 
     & \scriptsize \bf +CaP
     & \scriptsize \bf Ori   
     & \scriptsize \bf +CaP
     & \scriptsize \bf Ori 
     & \scriptsize \bf +CaP  
     & \scriptsize \bf Ori 
     & \scriptsize \bf +CaP    
     & \scriptsize \bf Ori 
     & \scriptsize \bf +CaP  \\ 
    \midrule
       
        \scriptsize \multirow{3}{*}{ViT~\cite{dosovitskiy2020image}} & \scriptsize MAE  & $16.59$  & $14.54$  & $32.03$  & $29.52$  & $18.69$ &$17.56$  & $13.59$  & $7.52$  & $17.32$  & $15.97$  \\ 
        & \scriptsize MSE & $11.40$  &$8.89$ &$36.11$  & $32.79$  & $9.93$   &$9.16$ & $6.21$ & $1.41$  & $23.40$ & $21.06$  \\ 

        & \scriptsize $\Delta$ &  \cellcolor{blue!25}$12.4\% \uparrow$   & \cellcolor{red!25}$22.1\% \uparrow$    & \cellcolor{blue!25}$7.8\% \uparrow$   & \cellcolor{red!25}$9.2\% \uparrow$    & \cellcolor{blue!25}$6.1\% \uparrow$   & \cellcolor{red!25}$7.7\% \uparrow$  & \cellcolor{blue!25}$44.7\% \uparrow$  & \cellcolor{red!25}$77.3\% \uparrow$  & \cellcolor{blue!25}$7.8\% \uparrow$   & \cellcolor{red!25}$10.1\%\uparrow$   \\ \hdashline[1pt/1pt]

        \scriptsize \multirow{3}{*}{Earthfarsser~\cite{wu2024earthfarsser}}  & \scriptsize MAE  & $14.57$  & $12.75$  & $23.56$  & $20.59$ & $15.23$  & $14.47$  & $2.03$  & $1.44$  & $17.15$ & $16.29$  \\ 
        
        & \scriptsize MSE   & $9.94$  & $7.83$    & $16.84$  & $14.07$    & $6.75$   & $6.01$ & $4.09$  & $2.24$  & $23.37$  & $21.94$ \\

        & \scriptsize $\Delta$ & \cellcolor{blue!25}$12.5\%\uparrow$   & \cellcolor{red!25}$21.2\%\uparrow$  & \cellcolor{blue!25}$12.6\%\uparrow$   & \cellcolor{red!25}$16.4\%\uparrow$   & \cellcolor{blue!25}$5.0\%\uparrow$   & \cellcolor{red!25}$10.9\%\uparrow$ & \cellcolor{blue!25}$29.1\%\uparrow$   & \cellcolor{red!25}$37.8\%\uparrow$   & \cellcolor{blue!25} $5.1\%\uparrow$  & \cellcolor{red!25}$6.1\%\uparrow$   \\ \hdashline[1pt/1pt]
        
        \scriptsize \multirow{3}{*}{Mmvp~\cite{zhong2023mmvp}}    & \scriptsize MAE   & $17.41$ & $16.17$  & $30.62$  & $27.57$  & $20.67$  & $17.21$ & $15.05$  & $11.02$   &$19.37$   &$18.16$   \\  
          & \scriptsize MSE  & $14.22$ & $12.29$  & $27.31$   & $22.37$  & $8.45$ & $7.26$  & $4.11$  & $2.32$   &$26.09$  &$24.97$      \\  
          & \scriptsize $\Delta$ & \cellcolor{blue!25}$7.1\%\uparrow$  & \cellcolor{red!25}$13.6\%\uparrow$ & \cellcolor{blue!25}$10.0\%\uparrow$   & \cellcolor{red!25}$18.1\%\uparrow$   & \cellcolor{blue!25}$16.7\%\uparrow$   & \cellcolor{red!25}$14.1\%\uparrow$  & \cellcolor{blue!25}$26.8\%\uparrow$   & \cellcolor{red!25}$43.6\%\uparrow$    & \cellcolor{blue!25}$6.2\%\uparrow$    & \cellcolor{red!25}$4.3\%\uparrow$   \\ \hdashline[1pt/1pt]

        \scriptsize \multirow{3}{*}{ConvLSTM~\cite{shi2015convolutional}}   & \scriptsize MAE  & $18.22$   & $16.21$   &  $22.77$   & $20.03$ & $20.51$   & $18.41$ & $5.43$  & $3.89$  & $22.22$   & $10.08$    \\ 
        & \scriptsize MSE & $16.79$   & $14.67$   & $27.37$  & $25.15$ & $12.12$   & $11.41$ & $0.64$ & $0.31$  & $28.64$  & $26.44$  \\ 
        & \scriptsize $\Delta$ & \cellcolor{blue!25}$13.4\%\uparrow$    & \cellcolor{red!25}$12.6\%\uparrow$   &\cellcolor{blue!25}$12.1\%\uparrow$     & \cellcolor{red!25}$8.1\%\uparrow$     & \cellcolor{blue!25}$10.2\%\uparrow$   & \cellcolor{red!25}$5.9\%\uparrow$  & \cellcolor{blue!25}$28.3\%\uparrow$ & \cellcolor{red!25}$51.6\%\uparrow$ & \cellcolor{blue!25}$9.6\%\uparrow$  & \cellcolor{red!25}$7.6\%\uparrow$ \\ \hdashline[1pt/1pt]
        
        \scriptsize \multirow{3}{*}{PredRNN-V2~\cite{wang2022predrnn}}   & \scriptsize MAE & $14.18$  & $13.05$ & $26.73$  & $23.64$   &$17.94$  & $16.26$ & $8.76$  & $7.98$ & $18.26$ & $16.14$ \\ 
          
        & \scriptsize MSE & $9.60$  & $7.89$ & $21.45$  & $19.11$ & $8.54$  & $7.73$ & $4.37$  & $4.18$ & $24.71$  & $23.12$  \\ 

        & \scriptsize $\Delta$ & \cellcolor{blue!25}$8.0\% \uparrow$   & \cellcolor{red!25}$16.6\% \uparrow$ & \cellcolor{blue!25}$11.6\% \uparrow$  &\cellcolor{red!25}$10.9\% \uparrow$   & \cellcolor{blue!25}$9.3\% \uparrow$   & \cellcolor{red!25}$9.4\% \uparrow$  & \cellcolor{blue!25}$8.9\% \uparrow$  & \cellcolor{red!25}$4.3\% \uparrow$  & \cellcolor{blue!25} $11.6\% \uparrow$ & \cellcolor{red!25}$6.5\% \uparrow$\\ \hdashline[1pt/1pt]
        
        \scriptsize \multirow{3}{*}{MAU~\cite{chang2021mau}}   & \scriptsize MAE  & $23.28$ & $20.96$    & $29.54$  & $27.82$    & $25.07$   & $24.14$ & $11.84$ & $9.97$ & $20.67$  & $18.65$  \\  
        
         & \scriptsize MSE & $20.46$  & $16.60$  & $30.19$  & $27.84$  & $15.43$   & $14.34$ & $5.28$   & $4.66$  & $30.89$  & $28.91$     \\  
         
         & \scriptsize $\Delta$ & \cellcolor{blue!25}$10.0\%\uparrow$   & \cellcolor{red!25}$18.9\%\uparrow$  & \cellcolor{blue!25}$5.9\%\uparrow$    & \cellcolor{red!25}$7.8\%\uparrow$     & \cellcolor{blue!25}$3.7\%\uparrow$    & \cellcolor{red!25}$7.1\%\uparrow$   & \cellcolor{blue!25}$15.8\%\uparrow$  & \cellcolor{red!25}$11.8\%\uparrow$   & \cellcolor{blue!25}$9.8\%\uparrow$   & \cellcolor{red!25}$6.4\%\uparrow$      \\ \hdashline[1pt/1pt]
         
        \scriptsize \multirow{3}{*}{SimVP~\cite{gao2022simvp}}  & \scriptsize MAE & $15.91$ & $13.45$  & $23.21$ &  $20.56$ & $15.48$ & $14.63$ & $2.12$ & $1.57$   & $17.01$  & $15.79$ \\  
        & \scriptsize MSE & $10.96$ & $8.21$  & $16.46$ & $13.91$ & $6.82$ & $6.21$ & $9.54$ & $5.03$  & $23.34$ & $22.11$ \\ 
        
        & \scriptsize $\Delta$ & \cellcolor{blue!25}$15.4\% \uparrow$   & \cellcolor{red!25}$25.1\%\uparrow$  & \cellcolor{blue!25}$11.4\%\uparrow$ & \cellcolor{red!25}$15.3\%\uparrow$  & \cellcolor{blue!25} $5.5\% \uparrow$  & \cellcolor{red!25}$8.9\% \uparrow$  &  \cellcolor{blue!25}$25.9\% \uparrow$ &  \cellcolor{red!25}$47.3\% \uparrow$  & \cellcolor{blue!25}$8.4\% \uparrow$ & \cellcolor{red!25}$5.3\% \uparrow $ \\  \midrule  
\end{tabularx}
\end{center}
\vspace{-2em}
\end{table}

\vspace{-0.4em}
\subsection{Performance in Data-Scarce Scenarios (RQ2)}
\vspace{-0.4em}


\begin{wrapfigure}{r}{0.4\textwidth}
\vspace{-3em}
  \begin{center}
    \includegraphics[width=0.4\textwidth]{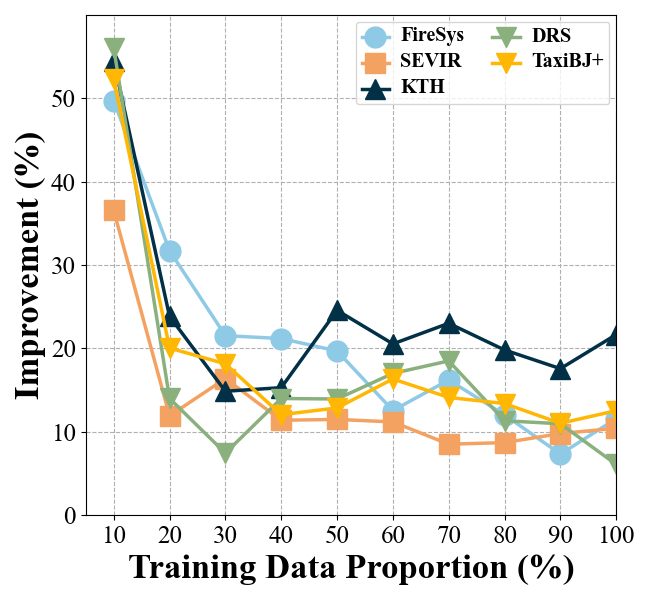}
  \end{center}
  \vspace{-1em}
  \caption{SSIM improvement across different datasets using the Mmvp model}
  \label{fig:improvement_using_Mmvp} 
  \vspace{-5em}
\end{wrapfigure}

To assess the performance of \textit{CaPaint} in data-scarce scenarios, we conducted experiments using \textbf{varying proportions of training data} across multiple datasets and backbones. Specifically, we measured the SSIM improvement at different training data proportions, demonstrating the generalizability and robustness of \textit{CaPaint}.

\begin{wrapfigure}{r}{0.4\textwidth}
\vspace{-1.5em}
  \begin{center}
    \includegraphics[width=0.4\textwidth]{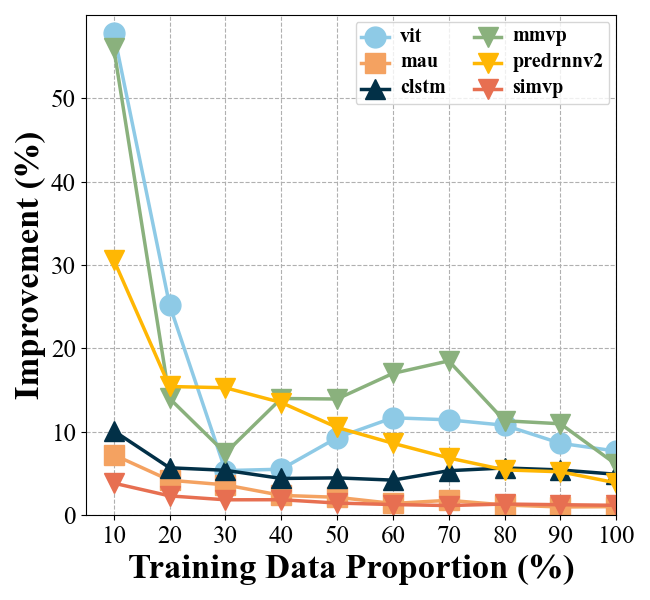}
  \end{center}
  \vspace{-1em}
  \caption{SSIM Improvement on DRS across various backbones}
  \label{fig:improvement_with_DRS} 
  \vspace{-1em}
\end{wrapfigure}

\noindent\textbf{Obs 1. CaPaint shows consistent improvements across all training data proportions.} As shown in Figures~\ref{fig:improvement_using_Mmvp} and ~\ref{fig:improvement_with_DRS}, \textit{CaPaint} consistently improves SSIM across all training data proportions. This indicates that CaPaint is effective regardless of the amount of training data available, reinforcing its versatility and applicability in diverse scenarios.

\noindent\textbf{Obs 2. Significant performance gains in low data scenarios.} The results indicate that CaPaint yields substantial performance improvements, especially in low data scenarios. For instance, with only 10\% of the training data, the SSIM improvement is most pronounced, highlighting the method's effectiveness in data-scarce environments. For example, in the TaxiBJ+ dataset with ViT backbone, the SSIM improvement reaches up to more than 50\%, showcasing CaPaint's capability to enhance model performance with limited data.

\noindent\textbf{Obs 3. Diminishing returns with increased training data.} While \textit{CaPaint} consistently enhances performance, the degree of improvement diminishes as the proportion of training data increases. This trend suggests that the primary benefits of \textit{CaPaint} are most evident when data is scarce, but the method remains beneficial even as more data becomes available.

\noindent\textbf{Obs 4. CaPaint demonstrates superior performance with equivalent data volumes.} As illustrated in Fig \ref{fig:abl1}, when comparing 25\% original plus 25\% augmented data with 50\% original data, \textit{CaPaint} achieves lower MAE and MSE. This demonstrates that \textit{CaPaint} consistently outperforms the original model by effectively using a mix of original and augmented data, which together match the data volume used by the original model alone.

\vspace{-0.4em}
\subsection{Performance Comparison (RQ3)}
\vspace{-0.4em}

\begin{minipage}[!t]{\linewidth}
    \begin{minipage}[!t]{0.60\linewidth}
    \centering
    \renewcommand{\arraystretch}{1.1}
    \scriptsize
    \tabcolsep=0.25mm
    \captionof{table}{Comparison between CaPaint and other data augmentation methods across various datasets.}
    \label{tab:data_augmentation_comparison}
    \begin{tabular}{ccccccc}
        \toprule
        Datasets & Flip & Rotate & Crop & NuWa & CaPaint \\
        \midrule
        DRS  & $2.10_{\pm 0.16}$ & $2.11_{\pm 0.19} $ & $2.34_{\pm 0.26}$ & $2.02_{\pm 0.09}$ & $1.57_{\pm 0.14}$ \\
        KTH  & $23.15_{\pm 1.95}$ & $23.14_{\pm 1.67}$ & $23.11_{\pm 1.83}$ & $22.32_{\pm 0.94}$ & $20.56_{\pm 1.02}$\\
        SEVIR  & $15.41_{\pm 1.49}$ & $15.45_{\pm 1.32}$ & $15.95_{\pm 1.64}$ & $15.14_{\pm 1.57}$ & $14.63_{\pm 1.89}$ \\
        TaxiBJ+  & $16.47_{\pm 0.99}$ & $16.39_{\pm 1.32}$& $15.94_{\pm 1.45}$ & $15.11_{\pm 0.87}$ & $12.87_{\pm 0.76}$ \\
        FireSys  & $17.02_{\pm 2.17}$ & $17.07_{\pm 1.94}$ & $17.15_{\pm 2.45}$ & $16.68_{\pm 1.79}$ & $15.79_{\pm 1.88}$ \\
        \bottomrule
      \end{tabular}
    \vspace{-0.5em} 

    \end{minipage}
    \hspace{0.1em}
    \begin{minipage}[!t]{0.40\linewidth}
    \centering
    \includegraphics[width=.95\linewidth]{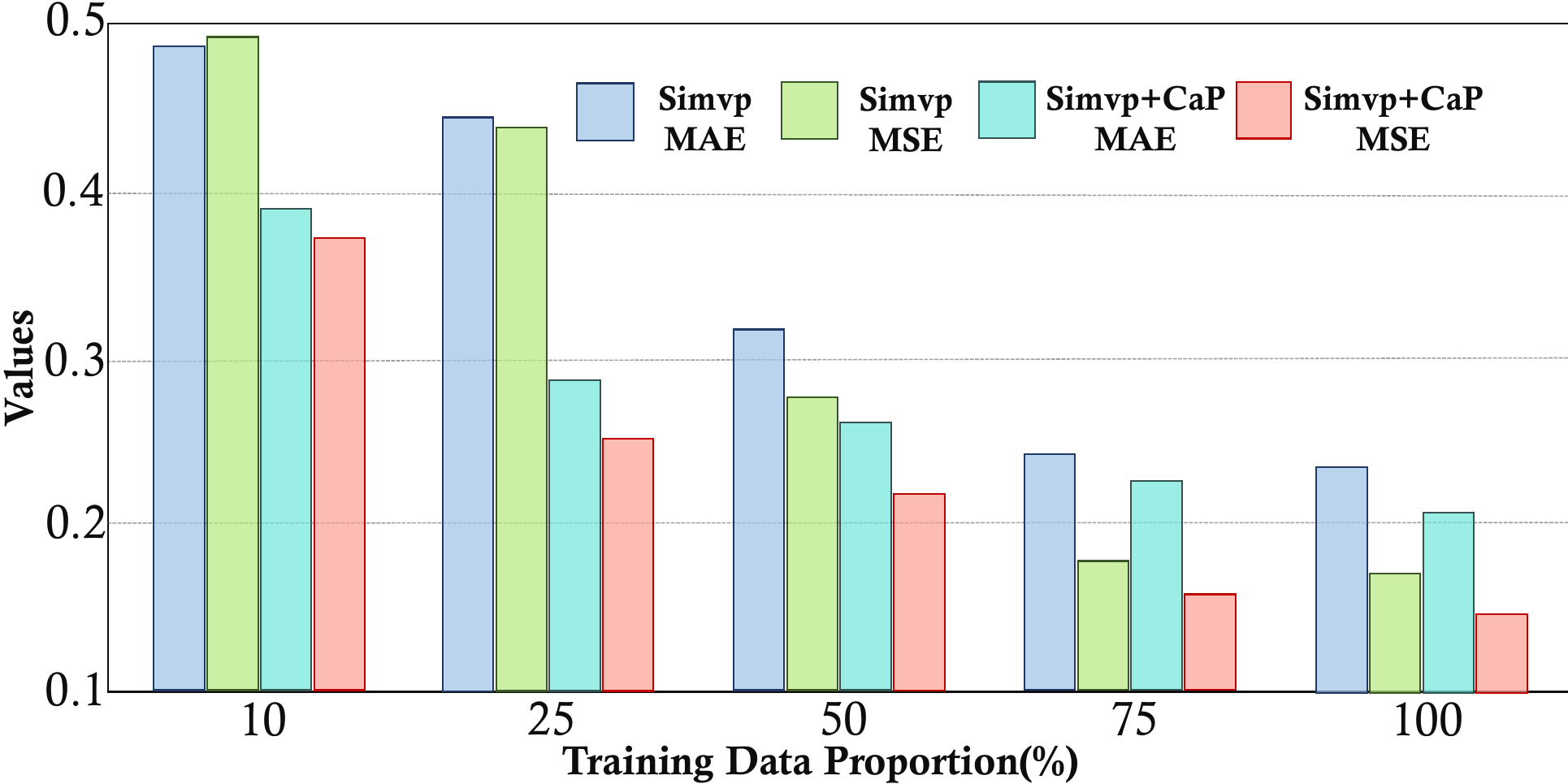}
 \vspace{-0.5em} \makeatletter\def\@captype{figure}\makeatother\caption{Visualizations in both MAE and MSE with Simvp and + CaP at various training data proportions. }\label{fig:abl1}
    \label{fig:bar}
    \end{minipage}
\end{minipage}

%

In this section, we compare the performance of different data augmentation methods. Tab \ref{tab:data_augmentation_comparison} shows the model performance using various data augmentation methods across multiple datasets, measured by MAE. It can be seen that traditional data augmentation methods, such as flipping, rotation, and cropping, produce results that are either on par with or slightly worse than the original data. Take the FireSys dataset as an example, MAE increased from 17.01 $\rightarrow$ 17.07 after rotation augmentation. This indicates that conventional data augmentation methods may \textbf{disrupt the intrinsic properties} of ST data, thereby negatively impacting model performance.

In contrast, our method \textit{CaPaint} achieves the best performance \textbf{across all datasets}. For instance, on the TaxiBJ+ dataset, the MAE with \textit{CaPaint} augmentation is $12.87$, which is significantly better than the MAE of $15.11$ with \textit{NuwaDynamics} manual mixup augmentation and the MAE of $15.94$ with other traditional augmentation methods such as cropping. These results highlight the advantage of our method in preserving the integrity of ST data properties. CaPaint not only effectively avoids the disruption caused by data augmentation processes on ST data characteristics but also significantly enhances the model's predictive capability.

\section{Conclusion \& Future Work}\label{sec:5}

In this study, we advance the exploration of applying front-door adjustment and causality principles to spatio-temporal forecasting tasks through the introduction of \textit{CaPaint}. Building upon the foundation of upstream self-supervised learning, we identify causal regions as crucial elements for generating comprehensive and potential data distributions. By integrating diffusion generative models, we ensure the generated data’s rationality and generalizability, thereby enhancing the downstream models’ ability to generalize beyond the observed distribution and improving their interpretability. Moving forward, we plan to explore various generative models for the production of arbitrary-channel ST data to enhance the \textit{CaPaint} robustness.

\section{Acknowledgement}\label{sec:6}
This work was supported by National Natural Science Foundation of China (62476224).

{
\bibliographystyle{plainnat}
\bibliography{reference}}

\begin{thebibliography}{106}
\providecommand{\natexlab}[1]{#1}
\providecommand{\url}[1]{\texttt{#1}}
\expandafter\ifx\csname urlstyle\endcsname\relax
  \providecommand{\doi}[1]{doi: #1}\else
  \providecommand{\doi}{doi: \begingroup \urlstyle{rm}\Url}\fi

\bibitem[Achiam et~al.(2023)Achiam, Adler, Agarwal, Ahmad, Akkaya, Aleman, Almeida, Altenschmidt, Altman, Anadkat, et~al.]{achiam2023gpt}
Josh Achiam, Steven Adler, Sandhini Agarwal, Lama Ahmad, Ilge Akkaya, Florencia~Leoni Aleman, Diogo Almeida, Janko Altenschmidt, Sam Altman, Shyamal Anadkat, et~al.
\newblock Gpt-4 technical report.
\newblock \emph{arXiv preprint arXiv:2303.08774}, 2023.

\bibitem[Bai et~al.(2022)Bai, Sun, Zhang, Song, and Chen]{bai2022rainformer}
Cong Bai, Feng Sun, Jinglin Zhang, Yi~Song, and Shengyong Chen.
\newblock Rainformer: Features extraction balanced network for radar-based precipitation nowcasting.
\newblock \emph{IEEE Geoscience and Remote Sensing Letters}, 19:\penalty0 1--5, 2022.

\bibitem[Bi et~al.(2022)Bi, Xie, Zhang, Chen, Gu, and Tian]{bi2022pangu}
Kaifeng Bi, Lingxi Xie, Hengheng Zhang, Xin Chen, Xiaotao Gu, and Qi~Tian.
\newblock Pangu-weather: A 3d high-resolution model for fast and accurate global weather forecast.
\newblock \emph{arXiv preprint arXiv:2211.02556}, 2022.

\bibitem[B{\"u}rkle et~al.(2021)B{\"u}rkle, Perera, Gimbert, Nakamura, Kawata, and Asai]{burkle2021deep}
Marius B{\"u}rkle, Umesha Perera, Florian Gimbert, Hisao Nakamura, Masaaki Kawata, and Yoshihiro Asai.
\newblock Deep-learning approach to first-principles transport simulations.
\newblock \emph{Physical Review Letters}, 126\penalty0 (17):\penalty0 177701, 2021.

\bibitem[Chang et~al.(2021)Chang, Zhang, Wang, Ma, Ye, Xinguang, and Gao]{chang2021mau}
Zheng Chang, Xinfeng Zhang, Shanshe Wang, Siwei Ma, Yan Ye, Xiang Xinguang, and Wen Gao.
\newblock Mau: A motion-aware unit for video prediction and beyond.
\newblock \emph{Advances in Neural Information Processing Systems}, 34:\penalty0 26950--26962, 2021.

\bibitem[Chen et~al.(2021)Chen, Huang, Raghupathi, Chandratreya, Du, and Lipson]{chen2021discovering}
Boyuan Chen, Kuang Huang, Sunand Raghupathi, Ishaan Chandratreya, Qiang Du, and Hod Lipson.
\newblock Discovering state variables hidden in experimental data.
\newblock \emph{arXiv preprint arXiv:2112.10755}, 2021.

\bibitem[Chen et~al.(2022)Chen, Huang, Raghupathi, Chandratreya, Du, and Lipson]{chen2022automated}
Boyuan Chen, Kuang Huang, Sunand Raghupathi, Ishaan Chandratreya, Qiang Du, and Hod Lipson.
\newblock Automated discovery of fundamental variables hidden in experimental data.
\newblock \emph{Nature Computational Science}, 2\penalty0 (7):\penalty0 433--442, 2022.

\bibitem[Chen and Dal~Negro(2022)]{chen2022physics}
Yuyao Chen and Luca Dal~Negro.
\newblock Physics-informed neural networks for imaging and parameter retrieval of photonic nanostructures from near-field data.
\newblock \emph{APL Photonics}, 7\penalty0 (1), 2022.

\bibitem[Cheng et~al.(2023)Cheng, Li, Liang, Sun, Yan, and Wu]{cheng2023rethinking}
Jinguo Cheng, Ke~Li, Yuxuan Liang, Lijun Sun, Junchi Yan, and Yuankai Wu.
\newblock Rethinking urban mobility prediction: A super-multivariate time series forecasting approach.
\newblock \emph{arXiv preprint arXiv:2312.01699}, 2023.

\bibitem[Cheng et~al.(2024)Cheng, Yang, Cai, Liang, and Wu]{cheng2024nuwats}
Jinguo Cheng, Chunwei Yang, Wanlin Cai, Yuxuan Liang, and Yuankai Wu.
\newblock Nuwats: Mending every incomplete time series.
\newblock \emph{arXiv preprint arXiv:2405.15317}, 2024.

\bibitem[Constantin and Foia{\c{s}}(1988)]{constantin1988navier}
Peter Constantin and Ciprian Foia{\c{s}}.
\newblock \emph{Navier-stokes equations}.
\newblock University of Chicago press, 1988.

\bibitem[Di~Capua et~al.(2020)Di~Capua, Runge, Donner, van~den Hurk, Turner, Vellore, Krishnan, and Coumou]{di2020dominant}
Giorgia Di~Capua, Jakob Runge, Reik~V Donner, Bart van~den Hurk, Andrew~G Turner, Ramesh Vellore, Raghavan Krishnan, and Dim Coumou.
\newblock Dominant patterns of interaction between the tropics and mid-latitudes in boreal summer: Causal relationships and the role of time-scales.
\newblock \emph{Weather and Climate Dynamics Discussions}, 2020:\penalty0 1--28, 2020.

\bibitem[Dosovitskiy et~al.(2020)Dosovitskiy, Beyer, Kolesnikov, Weissenborn, Zhai, Unterthiner, Dehghani, Minderer, Heigold, Gelly, et~al.]{dosovitskiy2020image}
Alexey Dosovitskiy, Lucas Beyer, Alexander Kolesnikov, Dirk Weissenborn, Xiaohua Zhai, Thomas Unterthiner, Mostafa Dehghani, Matthias Minderer, Georg Heigold, Sylvain Gelly, et~al.
\newblock An image is worth 16x16 words: Transformers for image recognition at scale.
\newblock \emph{arXiv preprint arXiv:2010.11929}, 2020.

\bibitem[Duan et~al.(2024)Duan, Zhang, Wang, Peng, Ziqi, Mao, Wu, Jiang, and Wang]{duan2024cat}
Yifan Duan, Guibin Zhang, Shilong Wang, Xiaojiang Peng, Wang Ziqi, Junyuan Mao, Hao Wu, Xinke Jiang, and Kun Wang.
\newblock Cat-gnn: Enhancing credit card fraud detection via causal temporal graph neural networks.
\newblock \emph{arXiv preprint arXiv:2402.14708}, 2024.

\bibitem[Ebert-Uphoff and Deng(2014)]{ebert2014causal}
Imme Ebert-Uphoff and Yi~Deng.
\newblock Causal discovery from spatio-temporal data with applications to climate science.
\newblock In \emph{2014 13th International Conference on Machine Learning and Applications}, pages 606--613. IEEE, 2014.

\bibitem[Fotiadis et~al.(2023)Fotiadis, Valencia, Hu, Garasto, Cantwell, and Bharath]{fotiadis2023disentangled}
Stathi Fotiadis, Mario~Lino Valencia, Shunlong Hu, Stef Garasto, Chris~D Cantwell, and Anil~Anthony Bharath.
\newblock Disentangled generative models for robust prediction of system dynamics.
\newblock 2023.

\bibitem[Gao et~al.(2023)Gao, Jiang, Zhuang, Chen, Wang, Law, and Haworth]{Gao2023UncertaintyAwarePG}
Xiaowei Gao, Xinke Jiang, Dingyi Zhuang, Huanfa Chen, Shenhao Wang, Stephen Law, and James Haworth.
\newblock Uncertainty-aware probabilistic graph neural networks for road-level traffic accident prediction.
\newblock 2023.
\newblock URL \url{https://api.semanticscholar.org/CorpusID:261681823}.

\bibitem[Gao et~al.(2022{\natexlab{a}})Gao, Tan, Wu, and Li]{gao2022simvp}
Zhangyang Gao, Cheng Tan, Lirong Wu, and Stan~Z Li.
\newblock Simvp: Simpler yet better video prediction.
\newblock In \emph{Proceedings of the IEEE/CVF conference on computer vision and pattern recognition}, pages 3170--3180, 2022{\natexlab{a}}.

\bibitem[Gao et~al.(2022{\natexlab{b}})Gao, Shi, Wang, Zhu, Wang, Li, and Yeung]{gao2022earthformer}
Zhihan Gao, Xingjian Shi, Hao Wang, Yi~Zhu, Yuyang~Bernie Wang, Mu~Li, and Dit-Yan Yeung.
\newblock Earthformer: Exploring space-time transformers for earth system forecasting.
\newblock \emph{Advances in Neural Information Processing Systems}, 35:\penalty0 25390--25403, 2022{\natexlab{b}}.

\bibitem[Gerard et~al.(2024)Gerard, Zhao, and Sullivan]{gerard2024wildfirespreadts}
Sebastian Gerard, Yu~Zhao, and Josephine Sullivan.
\newblock Wildfirespreadts: A dataset of multi-modal time series for wildfire spread prediction.
\newblock \emph{Advances in Neural Information Processing Systems}, 36, 2024.

\bibitem[Gulrajani and Lopez-Paz(2020)]{gulrajani2020search}
Ishaan Gulrajani and David Lopez-Paz.
\newblock In search of lost domain generalization.
\newblock \emph{arXiv preprint arXiv:2007.01434}, 2020.

\bibitem[Guo et~al.(2023)Guo, Zhang, Zhu, Tang, Ma, Han, Chen, Gao, Li, Li, et~al.]{guo2023point}
Ziyu Guo, Renrui Zhang, Xiangyang Zhu, Yiwen Tang, Xianzheng Ma, Jiaming Han, Kexin Chen, Peng Gao, Xianzhi Li, Hongsheng Li, et~al.
\newblock Point-bind \& point-llm: Aligning point cloud with multi-modality for 3d understanding, generation, and instruction following.
\newblock \emph{arXiv preprint arXiv:2309.00615}, 2023.

\bibitem[Han et~al.(2022)Han, Wang, Chen, Chen, Guo, Liu, Tang, Xiao, Xu, Xu, et~al.]{han2022survey}
Kai Han, Yunhe Wang, Hanting Chen, Xinghao Chen, Jianyuan Guo, Zhenhua Liu, Yehui Tang, An~Xiao, Chunjing Xu, Yixing Xu, et~al.
\newblock A survey on vision transformer.
\newblock \emph{IEEE transactions on pattern analysis and machine intelligence}, 45\penalty0 (1):\penalty0 87--110, 2022.

\bibitem[Ho et~al.(2020)Ho, Jain, and Abbeel]{ho2020denoising}
Jonathan Ho, Ajay Jain, and Pieter Abbeel.
\newblock Denoising diffusion probabilistic models.
\newblock \emph{Advances in neural information processing systems}, 33:\penalty0 6840--6851, 2020.

\bibitem[Jiang et~al.(2023)Jiang, Zhuang, Zhang, Chen, Luo, and Gao]{Jiang2023UncertaintyQV}
Xinke Jiang, Dingyi Zhuang, Xianghui Zhang, Hao Chen, Jiayuan Luo, and Xiaowei Gao.
\newblock Uncertainty quantification via spatial-temporal tweedie model for zero-inflated and long-tail travel demand prediction.
\newblock \emph{Proceedings of the 32nd ACM International Conference on Information and Knowledge Management}, 2023.
\newblock URL \url{https://api.semanticscholar.org/CorpusID:259187717}.

\bibitem[Jiang et~al.(2024)Jiang, Qin, Xu, and Ao]{Jiang2024IncompleteGL}
Xinke Jiang, Zidi Qin, Jiarong Xu, and Xiang Ao.
\newblock Incomplete graph learning via attribute-structure decoupled variational auto-encoder.
\newblock \emph{Proceedings of the 17th ACM International Conference on Web Search and Data Mining}, 2024.
\newblock URL \url{https://api.semanticscholar.org/CorpusID:268319406}.

\bibitem[Jin et~al.(2023{\natexlab{a}})Jin, Liang, Fang, Shao, Huang, Zhang, and Zheng]{jin2023spatio}
Guangyin Jin, Yuxuan Liang, Yuchen Fang, Zezhi Shao, Jincai Huang, Junbo Zhang, and Yu~Zheng.
\newblock Spatio-temporal graph neural networks for predictive learning in urban computing: A survey.
\newblock \emph{IEEE Transactions on Knowledge and Data Engineering}, 2023{\natexlab{a}}.

\bibitem[Jin et~al.(2023{\natexlab{b}})Jin, Wen, Liang, Zhang, Xue, Wang, Zhang, Wang, Chen, Li, et~al.]{jin2023large}
Ming Jin, Qingsong Wen, Yuxuan Liang, Chaoli Zhang, Siqiao Xue, Xue Wang, James Zhang, Yi~Wang, Haifeng Chen, Xiaoli Li, et~al.
\newblock Large models for time series and spatio-temporal data: A survey and outlook.
\newblock \emph{arXiv preprint arXiv:2310.10196}, 2023{\natexlab{b}}.

\bibitem[Kaffash et~al.(2021)Kaffash, Nguyen, and Zhu]{kaffash2021big}
Sepideh Kaffash, An~Truong Nguyen, and Joe Zhu.
\newblock Big data algorithms and applications in intelligent transportation system: A review and bibliometric analysis.
\newblock \emph{International journal of production economics}, 231:\penalty0 107868, 2021.

\bibitem[Karnewar et~al.(2023)Karnewar, Vedaldi, Novotny, and Mitra]{karnewar2023holodiffusion}
Animesh Karnewar, Andrea Vedaldi, David Novotny, and Niloy~J Mitra.
\newblock Holodiffusion: Training a 3d diffusion model using 2d images.
\newblock In \emph{Proceedings of the IEEE/CVF conference on computer vision and pattern recognition}, pages 18423--18433, 2023.

\bibitem[Karniadakis et~al.(2021)Karniadakis, Kevrekidis, Lu, Perdikaris, Wang, and Yang]{karniadakis2021physics}
George~Em Karniadakis, Ioannis~G Kevrekidis, Lu~Lu, Paris Perdikaris, Sifan Wang, and Liu Yang.
\newblock Physics-informed machine learning.
\newblock \emph{Nature Reviews Physics}, 3\penalty0 (6):\penalty0 422--440, 2021.

\bibitem[Karras et~al.(2019)Karras, Laine, and Aila]{karras2019style}
Tero Karras, Samuli Laine, and Timo Aila.
\newblock A style-based generator architecture for generative adversarial networks.
\newblock In \emph{Proceedings of the IEEE/CVF conference on computer vision and pattern recognition}, pages 4401--4410, 2019.

\bibitem[Khan et~al.(2022)Khan, Naseer, Hayat, Zamir, Khan, and Shah]{khan2022transformers}
Salman Khan, Muzammal Naseer, Munawar Hayat, Syed~Waqas Zamir, Fahad~Shahbaz Khan, and Mubarak Shah.
\newblock Transformers in vision: A survey.
\newblock \emph{ACM computing surveys (CSUR)}, 54\penalty0 (10s):\penalty0 1--41, 2022.

\bibitem[Koh et~al.(2021)Koh, Sagawa, Marklund, Xie, Zhang, Balsubramani, Hu, Yasunaga, Phillips, Gao, et~al.]{koh2021wilds}
Pang~Wei Koh, Shiori Sagawa, Henrik Marklund, Sang~Michael Xie, Marvin Zhang, Akshay Balsubramani, Weihua Hu, Michihiro Yasunaga, Richard~Lanas Phillips, Irena Gao, et~al.
\newblock Wilds: A benchmark of in-the-wild distribution shifts.
\newblock In \emph{International conference on machine learning}, pages 5637--5664. PMLR, 2021.

\bibitem[Krishnapriyan et~al.(2021)Krishnapriyan, Gholami, Zhe, Kirby, and Mahoney]{krishnapriyan2021characterizing}
Aditi Krishnapriyan, Amir Gholami, Shandian Zhe, Robert Kirby, and Michael~W Mahoney.
\newblock Characterizing possible failure modes in physics-informed neural networks.
\newblock \emph{Advances in Neural Information Processing Systems}, 34:\penalty0 26548--26560, 2021.

\bibitem[Li et~al.(2022)Li, Zhong, Jiang, Trajcevski, Wu, and Zhou]{Li2022MiningSR}
Rongfang Li, Ting Zhong, Xinke Jiang, Goce Trajcevski, Jin Wu, and Fan Zhou.
\newblock Mining spatio-temporal relations via self-paced graph contrastive learning.
\newblock \emph{Proceedings of the 28th ACM SIGKDD Conference on Knowledge Discovery and Data Mining}, 2022.
\newblock URL \url{https://api.semanticscholar.org/CorpusID:251518220}.

\bibitem[Liang et~al.(2021)Liang, Ouyang, Sun, Wang, Zhang, Zheng, Rosenblum, and Zimmermann]{liang2021fine}
Yuxuan Liang, Kun Ouyang, Junkai Sun, Yiwei Wang, Junbo Zhang, Yu~Zheng, David Rosenblum, and Roger Zimmermann.
\newblock Fine-grained urban flow prediction.
\newblock In \emph{Proceedings of the Web Conference 2021}, pages 1833--1845, 2021.

\bibitem[Liu et~al.(2011)Liu, Zheng, Chawla, Yuan, and Xing]{liu2011discovering}
Wei Liu, Yu~Zheng, Sanjay Chawla, Jing Yuan, and Xie Xing.
\newblock Discovering spatio-temporal causal interactions in traffic data streams.
\newblock In \emph{Proceedings of the 17th ACM SIGKDD international conference on Knowledge discovery and data mining}, pages 1010--1018, 2011.

\bibitem[Liu et~al.(2024)Liu, Xia, Liang, Hu, Wang, Bai, Huang, Liu, Hooi, and Zimmermann]{liu2024largest}
Xu~Liu, Yutong Xia, Yuxuan Liang, Junfeng Hu, Yiwei Wang, Lei Bai, Chao Huang, Zhenguang Liu, Bryan Hooi, and Roger Zimmermann.
\newblock Largest: A benchmark dataset for large-scale traffic forecasting.
\newblock \emph{Advances in Neural Information Processing Systems}, 36, 2024.

\bibitem[Long et~al.(2024)Long, Guo, Lin, Liu, Dou, Liu, Ma, Zhang, Habermann, Theobalt, et~al.]{long2024wonder3d}
Xiaoxiao Long, Yuan-Chen Guo, Cheng Lin, Yuan Liu, Zhiyang Dou, Lingjie Liu, Yuexin Ma, Song-Hai Zhang, Marc Habermann, Christian Theobalt, et~al.
\newblock Wonder3d: Single image to 3d using cross-domain diffusion.
\newblock In \emph{Proceedings of the IEEE/CVF Conference on Computer Vision and Pattern Recognition}, pages 9970--9980, 2024.

\bibitem[Lu et~al.(2022)Lu, Zhou, Bao, Chen, Li, and Zhu]{lu2022dpm}
Cheng Lu, Yuhao Zhou, Fan Bao, Jianfei Chen, Chongxuan Li, and Jun Zhu.
\newblock Dpm-solver++: Fast solver for guided sampling of diffusion probabilistic models.
\newblock \emph{arXiv preprint arXiv:2211.01095}, 2022.

\bibitem[Lugmayr et~al.(2022)Lugmayr, Danelljan, Romero, Yu, Timofte, and Van~Gool]{lugmayr2022repaint}
Andreas Lugmayr, Martin Danelljan, Andres Romero, Fisher Yu, Radu Timofte, and Luc Van~Gool.
\newblock Repaint: Inpainting using denoising diffusion probabilistic models.
\newblock In \emph{Proceedings of the IEEE/CVF conference on computer vision and pattern recognition}, pages 11461--11471, 2022.

\bibitem[Luo et~al.(2020)Luo, Cheng, Xu, Yu, Zong, Chen, and Zhang]{luo2020parameterized}
Dongsheng Luo, Wei Cheng, Dongkuan Xu, Wenchao Yu, Bo~Zong, Haifeng Chen, and Xiang Zhang.
\newblock Parameterized explainer for graph neural network.
\newblock \emph{Advances in neural information processing systems}, 33:\penalty0 19620--19631, 2020.

\bibitem[Luo et~al.(2024)Luo, Zhang, Fang, Gao, Zhuang, Chen, and Jiang]{Luo2024TimeseriesSA}
Jiayuan Luo, Wentao Zhang, Yuchen Fang, Xiaowei Gao, Dingyi Zhuang, Hao Chen, and Xinke Jiang.
\newblock Timeseries suppliers allocation risk optimization via deep black litterman model.
\newblock \emph{ArXiv}, abs/2401.17350, 2024.
\newblock URL \url{https://api.semanticscholar.org/CorpusID:271854629}.

\bibitem[Mathieu et~al.(2015)Mathieu, Couprie, and LeCun]{mathieu2015deep}
Michael Mathieu, Camille Couprie, and Yann LeCun.
\newblock Deep multi-scale video prediction beyond mean square error.
\newblock \emph{arXiv preprint arXiv:1511.05440}, 2015.

\bibitem[Nichol et~al.(2021)Nichol, Dhariwal, Ramesh, Shyam, Mishkin, McGrew, Sutskever, and Chen]{nichol2021glide}
Alex Nichol, Prafulla Dhariwal, Aditya Ramesh, Pranav Shyam, Pamela Mishkin, Bob McGrew, Ilya Sutskever, and Mark Chen.
\newblock Glide: Towards photorealistic image generation and editing with text-guided diffusion models.
\newblock \emph{arXiv preprint arXiv:2112.10741}, 2021.

\bibitem[Nowack et~al.(2020)Nowack, Runge, Eyring, and Haigh]{nowack2020causal}
Peer Nowack, Jakob Runge, Veronika Eyring, and Joanna~D Haigh.
\newblock Causal networks for climate model evaluation and constrained projections.
\newblock \emph{Nature communications}, 11\penalty0 (1):\penalty0 1415, 2020.

\bibitem[Oh et~al.(2015)Oh, Guo, Lee, Lewis, and Singh]{oh2015action}
Junhyuk Oh, Xiaoxiao Guo, Honglak Lee, Richard~L Lewis, and Satinder Singh.
\newblock Action-conditional video prediction using deep networks in atari games.
\newblock \emph{Advances in neural information processing systems}, 28, 2015.

\bibitem[Pathak et~al.(2016)Pathak, Krahenbuhl, Donahue, Darrell, and Efros]{pathak2016context}
Deepak Pathak, Philipp Krahenbuhl, Jeff Donahue, Trevor Darrell, and Alexei~A Efros.
\newblock Context encoders: Feature learning by inpainting.
\newblock In \emph{Proceedings of the IEEE conference on computer vision and pattern recognition}, pages 2536--2544, 2016.

\bibitem[Pathak et~al.(2022)Pathak, Subramanian, Harrington, Raja, Chattopadhyay, Mardani, Kurth, Hall, Li, Azizzadenesheli, et~al.]{pathak2022fourcastnet}
Jaideep Pathak, Shashank Subramanian, Peter Harrington, Sanjeev Raja, Ashesh Chattopadhyay, Morteza Mardani, Thorsten Kurth, David Hall, Zongyi Li, Kamyar Azizzadenesheli, et~al.
\newblock Fourcastnet: A global data-driven high-resolution weather model using adaptive fourier neural operators.
\newblock \emph{arXiv preprint arXiv:2202.11214}, 2022.

\bibitem[Pearl(2009)]{pearl2009causality}
Judea Pearl.
\newblock \emph{Causality}.
\newblock Cambridge university press, 2009.

\bibitem[Pearl and Mackenzie(2018)]{pearl2018book}
Judea Pearl and Dana Mackenzie.
\newblock \emph{The book of why: the new science of cause and effect}.
\newblock Basic books, 2018.

\bibitem[Pryor(2009)]{pryor2009multiphysics}
Roger~W Pryor.
\newblock \emph{Multiphysics modeling using COMSOL{\textregistered}: a first principles approach}.
\newblock Jones \& Bartlett Publishers, 2009.

\bibitem[Raissi et~al.(2019)Raissi, Perdikaris, and Karniadakis]{raissi2019physics}
Maziar Raissi, Paris Perdikaris, and George~E Karniadakis.
\newblock Physics-informed neural networks: A deep learning framework for solving forward and inverse problems involving nonlinear partial differential equations.
\newblock \emph{Journal of Computational physics}, 378:\penalty0 686--707, 2019.

\bibitem[Richardson et~al.(2021)Richardson, Alaluf, Patashnik, Nitzan, Azar, Shapiro, and Cohen-Or]{richardson2021encoding}
Elad Richardson, Yuval Alaluf, Or~Patashnik, Yotam Nitzan, Yaniv Azar, Stav Shapiro, and Daniel Cohen-Or.
\newblock Encoding in style: a stylegan encoder for image-to-image translation.
\newblock In \emph{Proceedings of the IEEE/CVF conference on computer vision and pattern recognition}, pages 2287--2296, 2021.

\bibitem[Sagawa et~al.(2019)Sagawa, Koh, Hashimoto, and Liang]{sagawa2019distributionally}
Shiori Sagawa, Pang~Wei Koh, Tatsunori~B Hashimoto, and Percy Liang.
\newblock Distributionally robust neural networks for group shifts: On the importance of regularization for worst-case generalization.
\newblock \emph{arXiv preprint arXiv:1911.08731}, 2019.

\bibitem[Saharia et~al.(2022)Saharia, Chan, Chang, Lee, Ho, Salimans, Fleet, and Norouzi]{saharia2022palette}
Chitwan Saharia, William Chan, Huiwen Chang, Chris Lee, Jonathan Ho, Tim Salimans, David Fleet, and Mohammad Norouzi.
\newblock Palette: Image-to-image diffusion models.
\newblock In \emph{ACM SIGGRAPH 2022 conference proceedings}, pages 1--10, 2022.

\bibitem[Schuldt et~al.(2004)Schuldt, Laptev, and Caputo]{schuldt2004recognizing}
Christian Schuldt, Ivan Laptev, and Barbara Caputo.
\newblock Recognizing human actions: a local svm approach.
\newblock In \emph{Proceedings of the 17th International Conference on Pattern Recognition, 2004. ICPR 2004.}, volume~3, pages 32--36. IEEE, 2004.

\bibitem[Schultz et~al.(2021)Schultz, Betancourt, Gong, Kleinert, Langguth, Leufen, Mozaffari, and Stadtler]{schultz2021can}
Martin~G Schultz, Clara Betancourt, Bing Gong, Felix Kleinert, Michael Langguth, Lukas~Hubert Leufen, Amirpasha Mozaffari, and Scarlet Stadtler.
\newblock Can deep learning beat numerical weather prediction?
\newblock \emph{Philosophical Transactions of the Royal Society A}, 379\penalty0 (2194):\penalty0 20200097, 2021.

\bibitem[Selvaraju et~al.(2016)Selvaraju, Das, Vedantam, Cogswell, Parikh, and Batra]{selvaraju2016grad}
Ramprasaath~R Selvaraju, Abhishek Das, Ramakrishna Vedantam, Michael Cogswell, Devi Parikh, and Dhruv Batra.
\newblock Grad-cam: Why did you say that?
\newblock \emph{arXiv preprint arXiv:1611.07450}, 2016.

\bibitem[Selvaraju et~al.(2017)Selvaraju, Cogswell, Das, Vedantam, Parikh, and Batra]{selvaraju2017grad}
Ramprasaath~R Selvaraju, Michael Cogswell, Abhishek Das, Ramakrishna Vedantam, Devi Parikh, and Dhruv Batra.
\newblock Grad-cam: Visual explanations from deep networks via gradient-based localization.
\newblock In \emph{Proceedings of the IEEE international conference on computer vision}, pages 618--626, 2017.

\bibitem[Shen et~al.(2023)Shen, Ye, Zhang, Wang, Han, and Wei]{shen2023advancing}
Fei Shen, Hu~Ye, Jun Zhang, Cong Wang, Xiao Han, and Yang Wei.
\newblock Advancing pose-guided image synthesis with progressive conditional diffusion models.
\newblock In \emph{The Twelfth International Conference on Learning Representations}, 2023.

\bibitem[Shen et~al.(2024{\natexlab{a}})Shen, Jiang, He, Ye, Wang, Du, Li, and Tang]{shen2024imagdressing}
Fei Shen, Xin Jiang, Xin He, Hu~Ye, Cong Wang, Xiaoyu Du, Zechao Li, and Jinghui Tang.
\newblock Imagdressing-v1: Customizable virtual dressing.
\newblock \emph{arXiv preprint arXiv:2407.12705}, 2024{\natexlab{a}}.

\bibitem[Shen et~al.(2024{\natexlab{b}})Shen, Ye, Liu, Zhang, Wang, Han, and Yang]{shen2024boosting}
Fei Shen, Hu~Ye, Sibo Liu, Jun Zhang, Cong Wang, Xiao Han, and Wei Yang.
\newblock Boosting consistency in story visualization with rich-contextual conditional diffusion models.
\newblock \emph{arXiv preprint arXiv:2407.02482}, 2024{\natexlab{b}}.

\bibitem[Shi et~al.(2015)Shi, Chen, Wang, Yeung, Wong, and Woo]{shi2015convolutional}
Xingjian Shi, Zhourong Chen, Hao Wang, Dit-Yan Yeung, Wai-Kin Wong, and Wang-chun Woo.
\newblock Convolutional lstm network: A machine learning approach for precipitation nowcasting.
\newblock \emph{Advances in neural information processing systems}, 28, 2015.

\bibitem[Shimizu et~al.(2006)Shimizu, Hoyer, Hyv{\"a}rinen, Kerminen, and Jordan]{shimizu2006linear}
Shohei Shimizu, Patrik~O Hoyer, Aapo Hyv{\"a}rinen, Antti Kerminen, and Michael Jordan.
\newblock A linear non-gaussian acyclic model for causal discovery.
\newblock \emph{Journal of Machine Learning Research}, 7\penalty0 (10), 2006.

\bibitem[Song et~al.(2020{\natexlab{a}})Song, Meng, and Ermon]{song2020denoising}
Jiaming Song, Chenlin Meng, and Stefano Ermon.
\newblock Denoising diffusion implicit models.
\newblock \emph{arXiv preprint arXiv:2010.02502}, 2020{\natexlab{a}}.

\bibitem[Song et~al.(2020{\natexlab{b}})Song, Sohl-Dickstein, Kingma, Kumar, Ermon, and Poole]{song2020score}
Yang Song, Jascha Sohl-Dickstein, Diederik~P Kingma, Abhishek Kumar, Stefano Ermon, and Ben Poole.
\newblock Score-based generative modeling through stochastic differential equations.
\newblock \emph{arXiv preprint arXiv:2011.13456}, 2020{\natexlab{b}}.

\bibitem[Srivastava et~al.(2015)Srivastava, Mansimov, and Salakhudinov]{srivastava2015unsupervised}
Nitish Srivastava, Elman Mansimov, and Ruslan Salakhudinov.
\newblock Unsupervised learning of video representations using lstms.
\newblock In \emph{International conference on machine learning}, pages 843--852. PMLR, 2015.

\bibitem[Takamoto et~al.(2022)Takamoto, Alesiani, and Niepert]{takamoto2022cape}
Makoto Takamoto, Francesco Alesiani, and Mathias Niepert.
\newblock Cape: Channel-attention-based pde parameter embeddings for sciml.
\newblock 2022.

\bibitem[Tam et~al.(2022)Tam, Fu, Li, Huang, Chen, and Huang]{tam2022spatial}
Wai~Cheong Tam, Eugene~Yujun Fu, Jiajia Li, Xinyan Huang, Jian Chen, and Michael~Xuelin Huang.
\newblock A spatial temporal graph neural network model for predicting flashover in arbitrary building floorplans.
\newblock \emph{Engineering Applications of Artificial Intelligence}, 115:\penalty0 105258, 2022.

\bibitem[Tan et~al.(2023)Tan, Gao, Wu, Xu, Xia, Li, and Li]{tan2023temporal}
Cheng Tan, Zhangyang Gao, Lirong Wu, Yongjie Xu, Jun Xia, Siyuan Li, and Stan~Z Li.
\newblock Temporal attention unit: Towards efficient spatiotemporal predictive learning.
\newblock In \emph{Proceedings of the IEEE/CVF Conference on Computer Vision and Pattern Recognition}, pages 18770--18782, 2023.

\bibitem[Tang et~al.(2024{\natexlab{a}})Tang, Liu, Wang, Wang, Zhang, Zhao, and Li]{tang2024any2point}
Yiwen Tang, Jiaming Liu, Dong Wang, Zhigang Wang, Shanghang Zhang, Bin Zhao, and Xuelong Li.
\newblock Any2point: Empowering any-modality large models for efficient 3d understanding.
\newblock \emph{arXiv preprint arXiv:2404.07989}, 2024{\natexlab{a}}.

\bibitem[Tang et~al.(2024{\natexlab{b}})Tang, Zhang, Guo, Ma, Zhao, Wang, Wang, and Li]{tang2024point}
Yiwen Tang, Ray Zhang, Zoey Guo, Xianzheng Ma, Bin Zhao, Zhigang Wang, Dong Wang, and Xuelong Li.
\newblock Point-peft: Parameter-efficient fine-tuning for 3d pre-trained models.
\newblock In \emph{Proceedings of the AAAI Conference on Artificial Intelligence}, volume~38, pages 5171--5179, 2024{\natexlab{b}}.

\bibitem[Tibau et~al.(2022)Tibau, Reimers, Gerhardus, Denzler, Eyring, and Runge]{tibau2022spatiotemporal}
Xavier-Andoni Tibau, Christian Reimers, Andreas Gerhardus, Joachim Denzler, Veronika Eyring, and Jakob Runge.
\newblock A spatiotemporal stochastic climate model for benchmarking causal discovery methods for teleconnections.
\newblock \emph{Environmental Data Science}, 1:\penalty0 e12, 2022.

\bibitem[Touvron et~al.(2023)Touvron, Lavril, Izacard, Martinet, Lachaux, Lacroix, Rozi{\`e}re, Goyal, Hambro, Azhar, et~al.]{touvron2023llama}
Hugo Touvron, Thibaut Lavril, Gautier Izacard, Xavier Martinet, Marie-Anne Lachaux, Timoth{\'e}e Lacroix, Baptiste Rozi{\`e}re, Naman Goyal, Eric Hambro, Faisal Azhar, et~al.
\newblock Llama: Open and efficient foundation language models.
\newblock \emph{arXiv preprint arXiv:2302.13971}, 2023.

\bibitem[Tulyakov et~al.(2018)Tulyakov, Liu, Yang, and Kautz]{tulyakov2018mocogan}
Sergey Tulyakov, Ming-Yu Liu, Xiaodong Yang, and Jan Kautz.
\newblock Mocogan: Decomposing motion and content for video generation.
\newblock In \emph{Proceedings of the IEEE conference on computer vision and pattern recognition}, pages 1526--1535, 2018.

\bibitem[Vaswani et~al.(2017)Vaswani, Shazeer, Parmar, Uszkoreit, Jones, Gomez, Kaiser, and Polosukhin]{vaswani2017attention}
Ashish Vaswani, Noam Shazeer, Niki Parmar, Jakob Uszkoreit, Llion Jones, Aidan~N Gomez, {\L}ukasz Kaiser, and Illia Polosukhin.
\newblock Attention is all you need.
\newblock \emph{Advances in neural information processing systems}, 30, 2017.

\bibitem[Veillette et~al.(2020)Veillette, Samsi, and Mattioli]{veillette2020sevir}
Mark Veillette, Siddharth Samsi, and Chris Mattioli.
\newblock Sevir: A storm event imagery dataset for deep learning applications in radar and satellite meteorology.
\newblock \emph{Advances in Neural Information Processing Systems}, 33:\penalty0 22009--22019, 2020.

\bibitem[Villegas et~al.(2018)Villegas, Erhan, Lee, et~al.]{villegas2018hierarchical}
Ruben Villegas, Dumitru Erhan, Honglak Lee, et~al.
\newblock Hierarchical long-term video prediction without supervision.
\newblock In \emph{International Conference on Machine Learning}, pages 6038--6046. PMLR, 2018.

\bibitem[Wang et~al.(2023{\natexlab{a}})Wang, Fu, Du, Gao, Huang, Liu, Chandak, Liu, Van~Katwyk, Deac, et~al.]{wang2023scientific}
Hanchen Wang, Tianfan Fu, Yuanqi Du, Wenhao Gao, Kexin Huang, Ziming Liu, Payal Chandak, Shengchao Liu, Peter Van~Katwyk, Andreea Deac, et~al.
\newblock Scientific discovery in the age of artificial intelligence.
\newblock \emph{Nature}, 620\penalty0 (7972):\penalty0 47--60, 2023{\natexlab{a}}.

\bibitem[Wang et~al.()Wang, Wu, Duan, Zhang, Wang, Peng, Zheng, Liang, and Wang]{wangnuwadynamics}
Kun Wang, Hao Wu, Yifan Duan, Guibin Zhang, Kai Wang, Xiaojiang Peng, Yu~Zheng, Yuxuan Liang, and Yang Wang.
\newblock Nuwadynamics: Discovering and updating in causal spatio-temporal modeling.

\bibitem[Wang et~al.(2024)Wang, Zhang, Zhang, Fang, Wu, Li, Pan, Huang, and Liang]{wang2024heterophilic}
Kun Wang, Guibin Zhang, Xinnan Zhang, Junfeng Fang, Xun Wu, Guohao Li, Shirui Pan, Wei Huang, and Yuxuan Liang.
\newblock The heterophilic snowflake hypothesis: Training and empowering gnns for heterophilic graphs.
\newblock In \emph{Proceedings of the 30th ACM SIGKDD Conference on Knowledge Discovery and Data Mining}, pages 3164--3175, 2024.

\bibitem[Wang et~al.(2020{\natexlab{a}})Wang, Cao, and Philip]{wang2020deep}
Senzhang Wang, Jiannong Cao, and S~Yu Philip.
\newblock Deep learning for spatio-temporal data mining: A survey.
\newblock \emph{IEEE transactions on knowledge and data engineering}, 34\penalty0 (8):\penalty0 3681--3700, 2020{\natexlab{a}}.

\bibitem[Wang et~al.(2020{\natexlab{b}})Wang, Yao, Kwok, and Ni]{wang2020generalizing}
Yaqing Wang, Quanming Yao, James~T Kwok, and Lionel~M Ni.
\newblock Generalizing from a few examples: A survey on few-shot learning.
\newblock \emph{ACM computing surveys (csur)}, 53\penalty0 (3):\penalty0 1--34, 2020{\natexlab{b}}.

\bibitem[Wang et~al.(2022{\natexlab{a}})Wang, Wu, Zhang, Gao, Wang, Philip, and Long]{wang2022predrnn}
Yunbo Wang, Haixu Wu, Jianjin Zhang, Zhifeng Gao, Jianmin Wang, S~Yu Philip, and Mingsheng Long.
\newblock Predrnn: A recurrent neural network for spatiotemporal predictive learning.
\newblock \emph{IEEE Transactions on Pattern Analysis and Machine Intelligence}, 45\penalty0 (2):\penalty0 2208--2225, 2022{\natexlab{a}}.

\bibitem[Wang et~al.(2022{\natexlab{b}})Wang, Sun, Hu, and Boukerche]{wang2022sfl}
Zepu Wang, Peng Sun, Yulin Hu, and Azzedine Boukerche.
\newblock Sfl: A high-precision traffic flow predictor for supporting intelligent transportation systems.
\newblock In \emph{GLOBECOM 2022-2022 IEEE Global Communications Conference}, pages 251--256. IEEE, 2022{\natexlab{b}}.

\bibitem[Wang et~al.(2023{\natexlab{b}})Wang, Sun, Lei, Zhu, and Sun]{wang2023sst}
Zepu Wang, Yifei Sun, Zhiyu Lei, Xincheng Zhu, and Peng Sun.
\newblock Sst: A simplified swin transformer-based model for taxi destination prediction based on existing trajectory.
\newblock In \emph{2023 IEEE 26th International Conference on Intelligent Transportation Systems (ITSC)}, pages 1404--1409. IEEE, 2023{\natexlab{b}}.

\bibitem[Wen et~al.(2022)Wen, Zhou, Zhang, Chen, Ma, Yan, and Sun]{wen2022transformers}
Qingsong Wen, Tian Zhou, Chaoli Zhang, Weiqi Chen, Ziqing Ma, Junchi Yan, and Liang Sun.
\newblock Transformers in time series: A survey.
\newblock \emph{arXiv preprint arXiv:2202.07125}, 2022.

\bibitem[Wu et~al.(2023{\natexlab{a}})Wu, Wang, Xu, Li, Wang, Wang, Wang, and Luo]{wu2023spatio}
Hao Wu, Kun Wang, Fan Xu, Yue Li, Xu~Wang, Weiyan Wang, Haixin Wang, and Xiao Luo.
\newblock Spatio-temporal twins with a cache for modeling long-term system dynamics.
\newblock 2023{\natexlab{a}}.

\bibitem[Wu et~al.(2023{\natexlab{b}})Wu, Xion, Xu, Luo, Chen, Hua, and Wang]{wu2023pastnet}
Hao Wu, Wei Xion, Fan Xu, Xiao Luo, Chong Chen, Xian-Sheng Hua, and Haixin Wang.
\newblock Pastnet: Introducing physical inductive biases for spatio-temporal video prediction.
\newblock \emph{arXiv preprint arXiv:2305.11421}, 2023{\natexlab{b}}.

\bibitem[Wu et~al.(2024{\natexlab{a}})Wu, Liang, Xiong, Zhou, Huang, Wang, and Wang]{wu2024earthfarsser}
Hao Wu, Yuxuan Liang, Wei Xiong, Zhengyang Zhou, Wei Huang, Shilong Wang, and Kun Wang.
\newblock Earthfarsser: Versatile spatio-temporal dynamical systems modeling in one model.
\newblock In \emph{Proceedings of the AAAI Conference on Artificial Intelligence}, volume~38, pages 15906--15914, 2024{\natexlab{a}}.

\bibitem[Wu et~al.(2024{\natexlab{b}})Wu, Wen, Zhang, Xia, Wang, Liang, Zheng, and Wang]{wu2024dynst}
Hao Wu, Haomin Wen, Guibin Zhang, Yutong Xia, Kai Wang, Yuxuan Liang, Yu~Zheng, and Kun Wang.
\newblock Dynst: Dynamic sparse training for resource-constrained spatio-temporal forecasting.
\newblock \emph{arXiv preprint arXiv:2403.02914}, 2024{\natexlab{b}}.

\bibitem[Wu et~al.(2022)Wu, Wang, Zhang, Hu, Feng, He, and Chua]{wu2022deconfounding}
Ying-Xin Wu, Xiang Wang, An~Zhang, Xia Hu, Fuli Feng, Xiangnan He, and Tat-Seng Chua.
\newblock Deconfounding to explanation evaluation in graph neural networks.
\newblock \emph{arXiv preprint arXiv:2201.08802}, 2022.

\bibitem[Xia et~al.(2024)Xia, Liang, Wen, Liu, Wang, Zhou, and Zimmermann]{xia2024deciphering}
Yutong Xia, Yuxuan Liang, Haomin Wen, Xu~Liu, Kun Wang, Zhengyang Zhou, and Roger Zimmermann.
\newblock Deciphering spatio-temporal graph forecasting: A causal lens and treatment.
\newblock \emph{Advances in Neural Information Processing Systems}, 36, 2024.

\bibitem[Yang et~al.(2024)Yang, Jiang, Zhao, Zeng, Liu, and Jia]{Yang2024FaiMAFI}
Songhua Yang, Xinke Jiang, Hanjie Zhao, Wenxuan Zeng, Hongde Liu, and Yuxiang Jia.
\newblock Faima: Feature-aware in-context learning for multi-domain aspect-based sentiment analysis.
\newblock \emph{ArXiv}, abs/2403.01063, 2024.
\newblock URL \url{https://api.semanticscholar.org/CorpusID:268230305}.

\bibitem[Ying et~al.(2019)Ying, Bourgeois, You, Zitnik, and Leskovec]{ying2019gnnexplainer}
Zhitao Ying, Dylan Bourgeois, Jiaxuan You, Marinka Zitnik, and Jure Leskovec.
\newblock Gnnexplainer: Generating explanations for graph neural networks.
\newblock \emph{Advances in neural information processing systems}, 32, 2019.

\bibitem[Zhang et~al.(2024{\natexlab{a}})Zhang, Wang, Huang, Yue, Wang, Zimmermann, Zhou, Cheng, Zeng, and Liang]{zhang2024graph}
Guibin Zhang, Kun Wang, Wei Huang, Yanwei Yue, Yang Wang, Roger Zimmermann, Aojun Zhou, Dawei Cheng, Jin Zeng, and Yuxuan Liang.
\newblock Graph lottery ticket automated.
\newblock In \emph{The Twelfth International Conference on Learning Representations}, 2024{\natexlab{a}}.

\bibitem[Zhang et~al.(2024{\natexlab{b}})Zhang, Yue, Wang, Fang, Sui, Wang, Liang, Cheng, Pan, and Chen]{zhang2024two}
Guibin Zhang, Yanwei Yue, Kun Wang, Junfeng Fang, Yongduo Sui, Kai Wang, Yuxuan Liang, Dawei Cheng, Shirui Pan, and Tianlong Chen.
\newblock Two heads are better than one: Boosting graph sparse training via semantic and topological awareness.
\newblock \emph{arXiv preprint arXiv:2402.01242}, 2024{\natexlab{b}}.

\bibitem[Zhang et~al.(2017)Zhang, Cisse, Dauphin, and Lopez-Paz]{zhang2017mixup}
Hongyi Zhang, Moustapha Cisse, Yann~N Dauphin, and David Lopez-Paz.
\newblock mixup: Beyond empirical risk minimization.
\newblock \emph{arXiv preprint arXiv:1710.09412}, 2017.

\bibitem[Zhao et~al.(2020)Zhao, Mo, Lin, Wang, Zuo, Chen, Xing, and Lu]{zhao2020uctgan}
Lei Zhao, Qihang Mo, Sihuan Lin, Zhizhong Wang, Zhiwen Zuo, Haibo Chen, Wei Xing, and Dongming Lu.
\newblock Uctgan: Diverse image inpainting based on unsupervised cross-space translation.
\newblock In \emph{Proceedings of the IEEE/CVF conference on computer vision and pattern recognition}, pages 5741--5750, 2020.

\bibitem[Zhao et~al.(2021)Zhao, Cui, Sheng, Dong, Liang, Chang, and Xu]{zhao2021large}
Shengyu Zhao, Jonathan Cui, Yilun Sheng, Yue Dong, Xiao Liang, Eric~I Chang, and Yan Xu.
\newblock Large scale image completion via co-modulated generative adversarial networks.
\newblock \emph{arXiv preprint arXiv:2103.10428}, 2021.

\bibitem[Zheng et~al.(2019)Zheng, Cham, and Cai]{zheng2019pluralistic}
Chuanxia Zheng, Tat-Jen Cham, and Jianfei Cai.
\newblock Pluralistic image completion.
\newblock In \emph{Proceedings of the IEEE/CVF Conference on Computer Vision and Pattern Recognition}, pages 1438--1447, 2019.

\bibitem[Zheng et~al.(2018)Zheng, Aragam, Ravikumar, and Xing]{zheng2018dags}
Xun Zheng, Bryon Aragam, Pradeep~K Ravikumar, and Eric~P Xing.
\newblock Dags with no tears: Continuous optimization for structure learning.
\newblock \emph{Advances in neural information processing systems}, 31, 2018.

\bibitem[Zhong et~al.(2023)Zhong, Liang, Zharkov, and Neumann]{zhong2023mmvp}
Yiqi Zhong, Luming Liang, Ilya Zharkov, and Ulrich Neumann.
\newblock Mmvp: Motion-matrix-based video prediction.
\newblock In \emph{Proceedings of the IEEE/CVF International Conference on Computer Vision}, pages 4273--4283, 2023.

\bibitem[Zhou et~al.(2023)Zhou, Huang, Yang, Wang, Wang, Zhang, Liang, and Wang]{zhou2023maintaining}
Zhengyang Zhou, Qihe Huang, Kuo Yang, Kun Wang, Xu~Wang, Yudong Zhang, Yuxuan Liang, and Yang Wang.
\newblock Maintaining the status quo: Capturing invariant relations for ood spatiotemporal learning.
\newblock In \emph{Proceedings of the 29th ACM SIGKDD Conference on Knowledge Discovery and Data Mining}, pages 3603--3614, 2023.

\end{thebibliography}


\begin{thebibliography}{65}
\providecommand{\natexlab}[1]{#1}
\providecommand{\url}[1]{\texttt{#1}}
\expandafter\ifx\csname urlstyle\endcsname\relax
  \providecommand{\doi}[1]{doi: #1}\else
  \providecommand{\doi}{doi: \begingroup \urlstyle{rm}\Url}\fi

\bibitem[Ahmadyan et~al.(2021)Ahmadyan, Zhang, Ablavatski, Wei, and
  Grundmann]{ahmadyan2021objectron}
Adel Ahmadyan, Liangkai Zhang, Artsiom Ablavatski, Jianing Wei, and Matthias
  Grundmann.
\newblock Objectron: A large scale dataset of object-centric videos in the wild
  with pose annotations.
\newblock In \emph{Proceedings of the IEEE/CVF Conference on Computer Vision
  and Pattern Recognition (CVPR)}, pages 7822--7831, 2021.

\bibitem[Alcorn et~al.(2019)Alcorn, Li, Gong, Wang, Mai, Ku, and
  Nguyen]{alcorn2019strike}
Michael~A Alcorn, Qi~Li, Zhitao Gong, Chengfei Wang, Long Mai, Wei-Shinn Ku,
  and Anh Nguyen.
\newblock Strike (with) a pose: Neural networks are easily fooled by strange
  poses of familiar objects.
\newblock In \emph{Proceedings of the IEEE/CVF Conference on Computer Vision
  and Pattern Recognition (CVPR)}, pages 4845--4854, 2019.

\bibitem[Athalye et~al.(2018)Athalye, Engstrom, Ilyas, and
  Kwok]{Athalye2017Synthesizing}
Anish Athalye, Logan Engstrom, Andrew Ilyas, and Kevin Kwok.
\newblock Synthesizing robust adversarial examples.
\newblock In \emph{International Conference on Machine Learning (ICML)}, pages
  284--293, 2018.

\bibitem[Bai et~al.(2021)Bai, Mei, Yuille, and Xie]{bai2021transformers}
Yutong Bai, Jieru Mei, Alan~L Yuille, and Cihang Xie.
\newblock Are transformers more robust than cnns?
\newblock In \emph{Advances in Neural Information Processing Systems
  (NeurIPS)}, pages 26831--26843, 2021.

\bibitem[Barbu et~al.(2019)Barbu, Mayo, Alverio, Luo, Wang, Gutfreund,
  Tenenbaum, and Katz]{barbu2019objectnet}
Andrei Barbu, David Mayo, Julian Alverio, William Luo, Christopher Wang, Dan
  Gutfreund, Josh Tenenbaum, and Boris Katz.
\newblock Objectnet: A large-scale bias-controlled dataset for pushing the
  limits of object recognition models.
\newblock In \emph{Advances in Neural Information Processing Systems
  (NeurIPS)}, pages 9453--9463, 2019.

\bibitem[Bhojanapalli et~al.(2021)Bhojanapalli, Chakrabarti, Glasner, Li,
  Unterthiner, and Veit]{bhojanapalli2021understanding}
Srinadh Bhojanapalli, Ayan Chakrabarti, Daniel Glasner, Daliang Li, Thomas
  Unterthiner, and Andreas Veit.
\newblock Understanding robustness of transformers for image classification.
\newblock In \emph{Proceedings of the IEEE/CVF International Conference on
  Computer Vision (ICCV)}, pages 10231--10241, 2021.

\bibitem[Biederman(1987)]{biederman1987recognition}
Irving Biederman.
\newblock Recognition-by-components: a theory of human image understanding.
\newblock \emph{Psychological review}, 94\penalty0 (2):\penalty0 115, 1987.

\bibitem[Blundell et~al.(2015)Blundell, Cornebise, Kavukcuoglu, and
  Wierstra]{blundell2015weight}
Charles Blundell, Julien Cornebise, Koray Kavukcuoglu, and Daan Wierstra.
\newblock Weight uncertainty in neural networks.
\newblock In \emph{International Conference on Machine Learning (ICML)}, pages
  1613--1622, 2015.

\bibitem[Chang et~al.(2015)Chang, Funkhouser, Guibas, Hanrahan, Huang, Li,
  Savarese, Savva, Song, Su, et~al.]{chang2015shapenet}
Angel~X Chang, Thomas Funkhouser, Leonidas Guibas, Pat Hanrahan, Qixing Huang,
  Zimo Li, Silvio Savarese, Manolis Savva, Shuran Song, Hao Su, et~al.
\newblock Shapenet: An information-rich 3d model repository.
\newblock \emph{arXiv preprint arXiv:1512.03012}, 2015.

\bibitem[Dong et~al.(2018)Dong, Liao, Pang, Su, Zhu, Hu, and Li]{Dong2017}
Yinpeng Dong, Fangzhou Liao, Tianyu Pang, Hang Su, Jun Zhu, Xiaolin Hu, and
  Jianguo Li.
\newblock Boosting adversarial attacks with momentum.
\newblock In \emph{Proceedings of the IEEE Conference on Computer Vision and
  Pattern Recognition (CVPR)}, pages 9185--9193, 2018.

\bibitem[Dong et~al.(2020)Dong, Deng, Pang, Zhu, and Su]{dong2020adversarial}
Yinpeng Dong, Zhijie Deng, Tianyu Pang, Jun Zhu, and Hang Su.
\newblock Adversarial distributional training for robust deep learning.
\newblock In \emph{Advances in Neural Information Processing Systems
  (NeurIPS)}, pages 8270--8283, 2020.

\bibitem[Dosovitskiy et~al.(2021)Dosovitskiy, Beyer, Kolesnikov, Weissenborn,
  Zhai, Unterthiner, Dehghani, Minderer, Heigold, Gelly,
  et~al.]{dosovitskiy2020image}
Alexey Dosovitskiy, Lucas Beyer, Alexander Kolesnikov, Dirk Weissenborn,
  Xiaohua Zhai, Thomas Unterthiner, Mostafa Dehghani, Matthias Minderer, Georg
  Heigold, Sylvain Gelly, et~al.
\newblock An image is worth 16x16 words: Transformers for image recognition at
  scale.
\newblock In \emph{International Conference on Learning Representations
  (ICLR)}, 2021.

\bibitem[Engstrom et~al.(2019)Engstrom, Tran, Tsipras, Schmidt, and
  Madry]{engstrom2019exploring}
Logan Engstrom, Brandon Tran, Dimitris Tsipras, Ludwig Schmidt, and Aleksander
  Madry.
\newblock Exploring the landscape of spatial robustness.
\newblock In \emph{International Conference on Machine Learning (ICML)}, pages
  1802--1811, 2019.

\bibitem[Eslami et~al.(2018)Eslami, Jimenez~Rezende, Besse, Viola, Morcos,
  Garnelo, Ruderman, Rusu, Danihelka, Gregor, et~al.]{eslami2018neural}
SM~Ali Eslami, Danilo Jimenez~Rezende, Frederic Besse, Fabio Viola, Ari~S
  Morcos, Marta Garnelo, Avraham Ruderman, Andrei~A Rusu, Ivo Danihelka, Karol
  Gregor, et~al.
\newblock Neural scene representation and rendering.
\newblock \emph{Science}, 360\penalty0 (6394):\penalty0 1204--1210, 2018.

\bibitem[Garbin et~al.(2021)Garbin, Kowalski, Johnson, Shotton, and
  Valentin]{garbin2021fastnerf}
Stephan~J Garbin, Marek Kowalski, Matthew Johnson, Jamie Shotton, and Julien
  Valentin.
\newblock Fastnerf: High-fidelity neural rendering at 200fps.
\newblock In \emph{Proceedings of the IEEE/CVF International Conference on
  Computer Vision (ICCV)}, pages 14346--14355, 2021.

\bibitem[Geirhos et~al.(2018)Geirhos, Temme, Rauber, Sch{\"u}tt, Bethge, and
  Wichmann]{geirhos2018generalisation}
Robert Geirhos, Carlos~RM Temme, Jonas Rauber, Heiko~H Sch{\"u}tt, Matthias
  Bethge, and Felix~A Wichmann.
\newblock Generalisation in humans and deep neural networks.
\newblock In \emph{Advances in Neural Information Processing Systems
  (NeurIPS)}, pages 7549--7561, 2018.

\bibitem[Geirhos et~al.(2019)Geirhos, Rubisch, Michaelis, Bethge, Wichmann, and
  Brendel]{geirhos2018imagenet}
Robert Geirhos, Patricia Rubisch, Claudio Michaelis, Matthias Bethge, Felix~A
  Wichmann, and Wieland Brendel.
\newblock Imagenet-trained cnns are biased towards texture; increasing shape
  bias improves accuracy and robustness.
\newblock In \emph{International Conference on Learning Representations
  (ICLR)}, 2019.

\bibitem[Geirhos et~al.(2021)Geirhos, Narayanappa, Mitzkus, Thieringer, Bethge,
  Wichmann, and Brendel]{geirhos2021partial}
Robert Geirhos, Kantharaju Narayanappa, Benjamin Mitzkus, Tizian Thieringer,
  Matthias Bethge, Felix~A Wichmann, and Wieland Brendel.
\newblock Partial success in closing the gap between human and machine vision.
\newblock In \emph{Advances in Neural Information Processing Systems
  (NeurIPS)}, pages 23885--23899, 2021.

\bibitem[Goodfellow et~al.(2015)Goodfellow, Shlens, and
  Szegedy]{goodfellow2014explaining}
Ian~J Goodfellow, Jonathon Shlens, and Christian Szegedy.
\newblock Explaining and harnessing adversarial examples.
\newblock In \emph{International Conference on Learning Representations
  (ICLR)}, 2015.

\bibitem[Gortler et~al.(1996)Gortler, Grzeszczuk, Szeliski, and
  Cohen]{gortler1996lumigraph}
Steven~J Gortler, Radek Grzeszczuk, Richard Szeliski, and Michael~F Cohen.
\newblock The lumigraph.
\newblock In \emph{Proceedings of the 23rd Annual Conference on Computer
  Graphics and Interactive Techniques (SIGGRAPH)}, pages 43--54, 1996.

\bibitem[He et~al.(2016)He, Zhang, Ren, and Sun]{he2016deep}
Kaiming He, Xiangyu Zhang, Shaoqing Ren, and Jian Sun.
\newblock Deep residual learning for image recognition.
\newblock In \emph{Proceedings of the IEEE Conference on Computer Vision and
  Pattern Recognition (CVPR)}, pages 770--778, 2016.

\bibitem[He et~al.(2022)He, Chen, Xie, Li, Doll{\'a}r, and
  Girshick]{he2021masked}
Kaiming He, Xinlei Chen, Saining Xie, Yanghao Li, Piotr Doll{\'a}r, and Ross
  Girshick.
\newblock Masked autoencoders are scalable vision learners.
\newblock In \emph{Proceedings of the IEEE/CVF Conference on Computer Vision
  and Pattern Recognition (CVPR)}, pages 16000--16009, 2022.

\bibitem[Hendrycks and Dietterich(2019)]{hendrycks2018benchmarking}
Dan Hendrycks and Thomas Dietterich.
\newblock Benchmarking neural network robustness to common corruptions and
  perturbations.
\newblock In \emph{International Conference on Learning Representations
  (ICLR)}, 2019.

\bibitem[Hendrycks et~al.(2020)Hendrycks, Mu, Cubuk, Zoph, Gilmer, and
  Lakshminarayanan]{hendrycks2019augmix}
Dan Hendrycks, Norman Mu, Ekin~Dogus Cubuk, Barret Zoph, Justin Gilmer, and
  Balaji Lakshminarayanan.
\newblock Augmix: A simple data processing method to improve robustness and
  uncertainty.
\newblock In \emph{International Conference on Learning Representations
  (ICLR)}, 2020.

\bibitem[Hendrycks et~al.(2021)Hendrycks, Basart, Mu, Kadavath, Wang, Dorundo,
  Desai, Zhu, Parajuli, Guo, et~al.]{hendrycks2021many}
Dan Hendrycks, Steven Basart, Norman Mu, Saurav Kadavath, Frank Wang, Evan
  Dorundo, Rahul Desai, Tyler Zhu, Samyak Parajuli, Mike Guo, et~al.
\newblock The many faces of robustness: A critical analysis of
  out-of-distribution generalization.
\newblock In \emph{Proceedings of the IEEE/CVF International Conference on
  Computer Vision (ICCV)}, pages 8340--8349, 2021.

\bibitem[Huang et~al.(2017)Huang, Liu, Van Der~Maaten, and
  Weinberger]{huang2017densely}
Gao Huang, Zhuang Liu, Laurens Van Der~Maaten, and Kilian~Q Weinberger.
\newblock Densely connected convolutional networks.
\newblock In \emph{Proceedings of the IEEE Conference on Computer Vision and
  Pattern Recognition (CVPR)}, pages 4700--4708, 2017.

\bibitem[Kanbak et~al.(2018)Kanbak, Moosavi-Dezfooli, and
  Frossard]{kanbak2018geometric}
Can Kanbak, Seyed-Mohsen Moosavi-Dezfooli, and Pascal Frossard.
\newblock Geometric robustness of deep networks: analysis and improvement.
\newblock In \emph{Proceedings of the IEEE Conference on Computer Vision and
  Pattern Recognition (CVPR)}, pages 4441--4449, 2018.

\bibitem[Kingma and Ba(2015)]{Kingma2014}
Diederik Kingma and Jimmy Ba.
\newblock Adam: A method for stochastic optimization.
\newblock In \emph{International Conference on Learning Representations
  (ICLR)}, 2015.

\bibitem[Kingma and Welling(2014)]{kingma2013auto}
Diederik~P Kingma and Max Welling.
\newblock Auto-encoding variational bayes.
\newblock In \emph{International Conference on Learning Representations
  (ICLR)}, 2014.

\bibitem[Kurakin et~al.(2017)Kurakin, Goodfellow, and Bengio]{Kurakin2016}
Alexey Kurakin, Ian Goodfellow, and Samy Bengio.
\newblock Adversarial examples in the physical world.
\newblock In \emph{International Conference on Learning Representations (ICLR)
  Workshops}, 2017.

\bibitem[Laidlaw and Feizi(2019)]{laidlaw2019functional}
Cassidy Laidlaw and Soheil Feizi.
\newblock Functional adversarial attacks.
\newblock In \emph{Advances in Neural Information Processing Systems
  (NeurIPS)}, pages 10408--10418, 2019.

\bibitem[Levoy and Hanrahan(1996)]{levoy1996light}
Marc Levoy and Pat Hanrahan.
\newblock Light field rendering.
\newblock In \emph{Proceedings of the 23rd Annual Conference on Computer
  Graphics and Interactive Techniques (SIGGRAPH)}, pages 31--42, 1996.

\bibitem[Liao et~al.(2018)Liao, Liang, Dong, Pang, Hu, and
  Zhu]{liao2018defense}
Fangzhou Liao, Ming Liang, Yinpeng Dong, Tianyu Pang, Xiaolin Hu, and Jun Zhu.
\newblock Defense against adversarial attacks using high-level representation
  guided denoiser.
\newblock In \emph{Proceedings of the IEEE Conference on Computer Vision and
  Pattern Recognition (CVPR)}, pages 1778--1787, 2018.

\bibitem[Liu et~al.(2020)Liu, Gu, Zaw~Lin, Chua, and Theobalt]{liu2020neural}
Lingjie Liu, Jiatao Gu, Kyaw Zaw~Lin, Tat-Seng Chua, and Christian Theobalt.
\newblock Neural sparse voxel fields.
\newblock In \emph{Advances in Neural Information Processing Systems
  (NeurIPS)}, pages 15651--15663, 2020.

\bibitem[Liu et~al.(2021)Liu, Lin, Cao, Hu, Wei, Zhang, Lin, and
  Guo]{liu2021swin}
Ze~Liu, Yutong Lin, Yue Cao, Han Hu, Yixuan Wei, Zheng Zhang, Stephen Lin, and
  Baining Guo.
\newblock Swin transformer: Hierarchical vision transformer using shifted
  windows.
\newblock In \emph{Proceedings of the IEEE/CVF International Conference on
  Computer Vision (ICCV)}, pages 10012--10022, 2021.

\bibitem[Lombardi et~al.(2019)Lombardi, Simon, Saragih, Schwartz, Lehrmann, and
  Sheikh]{lombardi2019neural}
Stephen Lombardi, Tomas Simon, Jason Saragih, Gabriel Schwartz, Andreas
  Lehrmann, and Yaser Sheikh.
\newblock Neural volumes: learning dynamic renderable volumes from images.
\newblock \emph{ACM Transactions on Graphics (TOG)}, 38\penalty0 (4):\penalty0
  1--14, 2019.

\bibitem[Louizos and Welling(2017)]{louizos2017multiplicative}
Christos Louizos and Max Welling.
\newblock Multiplicative normalizing flows for variational bayesian neural
  networks.
\newblock In \emph{International Conference on Machine Learning (ICML)}, pages
  2218--2227, 2017.

\bibitem[Madry et~al.(2018)Madry, Makelov, Schmidt, Tsipras, and
  Vladu]{madry2017towards}
Aleksander Madry, Aleksandar Makelov, Ludwig Schmidt, Dimitris Tsipras, and
  Adrian Vladu.
\newblock Towards deep learning models resistant to adversarial attacks.
\newblock In \emph{International Conference on Learning Representations
  (ICLR)}, 2018.

\bibitem[Martin-Brualla et~al.(2021)Martin-Brualla, Radwan, Sajjadi, Barron,
  Dosovitskiy, and Duckworth]{martin2021nerf}
Ricardo Martin-Brualla, Noha Radwan, Mehdi~SM Sajjadi, Jonathan~T Barron,
  Alexey Dosovitskiy, and Daniel Duckworth.
\newblock Nerf in the wild: Neural radiance fields for unconstrained photo
  collections.
\newblock In \emph{Proceedings of the IEEE/CVF Conference on Computer Vision
  and Pattern Recognition (CVPR)}, pages 7210--7219, 2021.

\bibitem[Mescheder et~al.(2019)Mescheder, Oechsle, Niemeyer, Nowozin, and
  Geiger]{mescheder2019occupancy}
Lars Mescheder, Michael Oechsle, Michael Niemeyer, Sebastian Nowozin, and
  Andreas Geiger.
\newblock Occupancy networks: Learning 3d reconstruction in function space.
\newblock In \emph{Proceedings of the IEEE/CVF Conference on Computer Vision
  and Pattern Recognition (CVPR)}, pages 4460--4470, 2019.

\bibitem[Mildenhall et~al.(2020)Mildenhall, Srinivasan, Tancik, Barron,
  Ramamoorthi, and Ng]{mildenhall2020nerf}
Ben Mildenhall, Pratul~P Srinivasan, Matthew Tancik, Jonathan~T Barron, Ravi
  Ramamoorthi, and Ren Ng.
\newblock Nerf: Representing scenes as neural radiance fields for view
  synthesis.
\newblock In \emph{European Conference on Computer Vision (ECCV)}, pages
  405--421, 2020.

\bibitem[Naseer et~al.(2021)Naseer, Ranasinghe, Khan, Hayat, Shahbaz~Khan, and
  Yang]{naseer2021intriguing}
Muhammad~Muzammal Naseer, Kanchana Ranasinghe, Salman~H Khan, Munawar Hayat,
  Fahad Shahbaz~Khan, and Ming-Hsuan Yang.
\newblock Intriguing properties of vision transformers.
\newblock In \emph{Advances in Neural Information Processing Systems
  (NeurIPS)}, pages 23296--23308, 2021.

\bibitem[Park et~al.(2019)Park, Florence, Straub, Newcombe, and
  Lovegrove]{park2019deepsdf}
Jeong~Joon Park, Peter Florence, Julian Straub, Richard Newcombe, and Steven
  Lovegrove.
\newblock Deepsdf: Learning continuous signed distance functions for shape
  representation.
\newblock In \emph{Proceedings of the IEEE/CVF Conference on Computer Vision
  and Pattern Recognition (CVPR)}, pages 165--174, 2019.

\bibitem[Recht et~al.(2019)Recht, Roelofs, Schmidt, and
  Shankar]{recht2019imagenet}
Benjamin Recht, Rebecca Roelofs, Ludwig Schmidt, and Vaishaal Shankar.
\newblock Do imagenet classifiers generalize to imagenet?
\newblock In \emph{International Conference on Machine Learning (ICML)}, pages
  5389--5400, 2019.

\bibitem[Russakovsky et~al.(2015)Russakovsky, Deng, Su, Krause, Satheesh, Ma,
  Huang, Karpathy, Khosla, Bernstein, et~al.]{russakovsky2015imagenet}
Olga Russakovsky, Jia Deng, Hao Su, Jonathan Krause, Sanjeev Satheesh, Sean Ma,
  Zhiheng Huang, Andrej Karpathy, Aditya Khosla, Michael Bernstein, et~al.
\newblock Imagenet large scale visual recognition challenge.
\newblock \emph{International Journal of Computer Vision}, 115\penalty0
  (3):\penalty0 211--252, 2015.

\bibitem[Salman et~al.(2020)Salman, Ilyas, Engstrom, Kapoor, and
  Madry]{salman2020adversarially}
Hadi Salman, Andrew Ilyas, Logan Engstrom, Ashish Kapoor, and Aleksander Madry.
\newblock Do adversarially robust imagenet models transfer better?
\newblock In \emph{Advances in Neural Information Processing Systems
  (NeurIPS)}, pages 3533--3545, 2020.

\bibitem[Sandler et~al.(2018)Sandler, Howard, Zhu, Zhmoginov, and
  Chen]{sandler2018mobilenetv2}
Mark Sandler, Andrew Howard, Menglong Zhu, Andrey Zhmoginov, and Liang-Chieh
  Chen.
\newblock Mobilenetv2: Inverted residuals and linear bottlenecks.
\newblock In \emph{Proceedings of the IEEE Conference on Computer Vision and
  Pattern Recognition (CVPR)}, pages 4510--4520, 2018.

\bibitem[Schonberger and Frahm(2016)]{schonberger2016structure}
Johannes~L Schonberger and Jan-Michael Frahm.
\newblock Structure-from-motion revisited.
\newblock In \emph{Proceedings of the IEEE Conference on Computer Vision and
  Pattern Recognition (CVPR)}, pages 4104--4113, 2016.

\bibitem[Simonyan and Zisserman(2015)]{simonyan2014very}
Karen Simonyan and Andrew Zisserman.
\newblock Very deep convolutional networks for large-scale image recognition.
\newblock In \emph{International Conference on Learning Representations
  (ICLR)}, 2015.

\bibitem[Sitzmann et~al.(2019)Sitzmann, Zollh{\"o}fer, and
  Wetzstein]{sitzmann2019scene}
Vincent Sitzmann, Michael Zollh{\"o}fer, and Gordon Wetzstein.
\newblock Scene representation networks: Continuous 3d-structure-aware neural
  scene representations.
\newblock In \emph{Advances in Neural Information Processing Systems
  (NeurIPS)}, pages 1121--1132, 2019.

\bibitem[Sitzmann et~al.(2020)Sitzmann, Martel, Bergman, Lindell, and
  Wetzstein]{sitzmann2020implicit}
Vincent Sitzmann, Julien Martel, Alexander Bergman, David Lindell, and Gordon
  Wetzstein.
\newblock Implicit neural representations with periodic activation functions.
\newblock In \emph{Advances in Neural Information Processing Systems
  (NeurIPS)}, pages 7462--7473, 2020.

\bibitem[Sohl-Dickstein et~al.(2015)Sohl-Dickstein, Weiss, Maheswaranathan, and
  Ganguli]{sohl2015deep}
Jascha Sohl-Dickstein, Eric Weiss, Niru Maheswaranathan, and Surya Ganguli.
\newblock Deep unsupervised learning using nonequilibrium thermodynamics.
\newblock In \emph{International Conference on Machine Learning (ICML)}, pages
  2256--2265, 2015.

\bibitem[Szegedy et~al.(2014)Szegedy, Zaremba, Sutskever, Bruna, Erhan,
  Goodfellow, and Fergus]{szegedy2013intriguing}
Christian Szegedy, Wojciech Zaremba, Ilya Sutskever, Joan Bruna, Dumitru Erhan,
  Ian Goodfellow, and Rob Fergus.
\newblock Intriguing properties of neural networks.
\newblock In \emph{International Conference on Learning Representations
  (ICLR)}, 2014.

\bibitem[Szegedy et~al.(2016)Szegedy, Vanhoucke, Ioffe, Shlens, and
  Wojna]{szegedy2016rethinking}
Christian Szegedy, Vincent Vanhoucke, Sergey Ioffe, Jon Shlens, and Zbigniew
  Wojna.
\newblock Rethinking the inception architecture for computer vision.
\newblock In \emph{Proceedings of the IEEE Conference on Computer Vision and
  Pattern Recognition (CVPR)}, pages 2818--2826, 2016.

\bibitem[Szegedy et~al.(2017)Szegedy, Ioffe, Vanhoucke, and
  Alemi]{szegedy2017inception}
Christian Szegedy, Sergey Ioffe, Vincent Vanhoucke, and Alexander~A Alemi.
\newblock Inception-v4, inception-resnet and the impact of residual connections
  on learning.
\newblock In \emph{Proceedings of the Thirty-First AAAI Conference on
  Artificial Intelligence (AAAI)}, pages 4278--4284, 2017.

\bibitem[Tan and Le(2019)]{tan2019efficientnet}
Mingxing Tan and Quoc Le.
\newblock Efficientnet: Rethinking model scaling for convolutional neural
  networks.
\newblock In \emph{International Conference on Machine Learning (ICML)}, pages
  6105--6114, 2019.

\bibitem[Tolstikhin et~al.(2021)Tolstikhin, Houlsby, Kolesnikov, Beyer, Zhai,
  Unterthiner, Yung, Steiner, Keysers, Uszkoreit, et~al.]{tolstikhin2021mlp}
Ilya~O Tolstikhin, Neil Houlsby, Alexander Kolesnikov, Lucas Beyer, Xiaohua
  Zhai, Thomas Unterthiner, Jessica Yung, Andreas Steiner, Daniel Keysers,
  Jakob Uszkoreit, et~al.
\newblock Mlp-mixer: An all-mlp architecture for vision.
\newblock In \emph{Advances in Neural Information Processing Systems
  (NeurIPS)}, pages 24261--24272, 2021.

\bibitem[Torralba and Efros(2011)]{torralba2011unbiased}
A~Torralba and AA~Efros.
\newblock Unbiased look at dataset bias.
\newblock In \emph{Proceedings of the IEEE Conference on Computer Vision and
  Pattern Recognition (CVPR)}, pages 1521--1528, 2011.

\bibitem[Touvron et~al.(2021)Touvron, Cord, Douze, Massa, Sablayrolles, and
  J{\'e}gou]{touvron2021training}
Hugo Touvron, Matthieu Cord, Matthijs Douze, Francisco Massa, Alexandre
  Sablayrolles, and Herv{\'e} J{\'e}gou.
\newblock Training data-efficient image transformers \& distillation through
  attention.
\newblock In \emph{International Conference on Machine Learning (ICML)}, pages
  10347--10357, 2021.

\bibitem[Wierstra et~al.(2014)Wierstra, Schaul, Glasmachers, Sun, Peters, and
  Schmidhuber]{wierstra2014natural}
Daan Wierstra, Tom Schaul, Tobias Glasmachers, Yi~Sun, Jan Peters, and
  J{\"u}rgen Schmidhuber.
\newblock Natural evolution strategies.
\newblock \emph{Journal of Machine Learning Research}, 15\penalty0
  (1):\penalty0 949--980, 2014.

\bibitem[Wong et~al.(2019)Wong, Schmidt, and Kolter]{wong2019wasserstein}
Eric Wong, Frank Schmidt, and Zico Kolter.
\newblock Wasserstein adversarial examples via projected sinkhorn iterations.
\newblock In \emph{International Conference on Machine Learning (ICML)}, pages
  6808--6817, 2019.

\bibitem[Xiao et~al.(2018)Xiao, Zhu, Li, He, Liu, and Song]{xiao2018spatially}
Chaowei Xiao, Jun-Yan Zhu, Bo~Li, Warren He, Mingyan Liu, and Dawn Song.
\newblock Spatially transformed adversarial examples.
\newblock In \emph{International Conference on Learning Representations
  (ICLR)}, 2018.

\bibitem[Xie et~al.(2020)Xie, Tan, Gong, Wang, Yuille, and
  Le]{xie2020adversarial}
Cihang Xie, Mingxing Tan, Boqing Gong, Jiang Wang, Alan~L Yuille, and Quoc~V
  Le.
\newblock Adversarial examples improve image recognition.
\newblock In \emph{Proceedings of the IEEE/CVF Conference on Computer Vision
  and Pattern Recognition (CVPR)}, pages 819--828, 2020.

\bibitem[Zeng et~al.(2019)Zeng, Liu, Wang, Qiu, Xie, Tai, Tang, and
  Yuille]{zeng2019adversarial}
Xiaohui Zeng, Chenxi Liu, Yu-Siang Wang, Weichao Qiu, Lingxi Xie, Yu-Wing Tai,
  Chi-Keung Tang, and Alan~L Yuille.
\newblock Adversarial attacks beyond the image space.
\newblock In \emph{Proceedings of the IEEE/CVF Conference on Computer Vision
  and Pattern Recognition (CVPR)}, pages 4302--4311, 2019.

\bibitem[Zhang et~al.(2019)Zhang, Yu, Jiao, Xing, Ghaoui, and
  Jordan]{zhang2019theoretically}
Hongyang Zhang, Yaodong Yu, Jiantao Jiao, Eric~P Xing, Laurent~El Ghaoui, and
  Michael~I Jordan.
\newblock Theoretically principled trade-off between robustness and accuracy.
\newblock In \emph{International Conference on Machine Learning (ICML)}, pages
  7472--7482, 2019.

\end{thebibliography}

\clearpage
\appendix

\numberwithin{equation}{section}
\numberwithin{figure}{section}
\numberwithin{table}{section}

\section{CaPaint Inpainting Algorithm}\label{app:a}

\begin{center}
\begin{algorithm}
\caption{Causal Intervention with Diffusion Inpainting}
\label{alg:inpainting}
\begin{algorithmic}[1]
\State \textbf{Input:} ST observation data \( X \), masked image \( X_{mask} \)
\State \textbf{Output:} Augmentation ST observation dataset \( X_A \)
\vspace{0.5mm}
\State Initialize \( X_T \sim \mathcal{N}(0, I) \) where \( T \) is the total number of diffusion steps
\State \textcolor{blue}{/* Iterate backwards through diffusion steps */}
\For{\( t = T \) \textbf{to} 1}
    \State \textcolor{blue}{/* Sample Gaussian noise $\epsilon$ */}
    \State  $\epsilon \sim \mathcal{N}(0, I)$
    \State \textcolor{blue}{/* Sample causal region */}
    \State $X_{t-1}^{cau} = \sqrt{\bar\alpha_t} X_{0} + (1 - \bar\alpha_t) \epsilon $
    \vspace{0.5mm}
    \State \textcolor{blue}{/* Sample Gaussian noise $\mathcal{N}$ */}
    \State $ z \sim \mathcal{N}(0, I)$
    \State \textcolor{blue}{/* Causal Intervention on Environmental Patches */}
    \State 
    $
    X_{t-1}^{env} = \frac{1}{\sqrt{\alpha_t}} \left(X_t - \frac{\beta_t}{\sqrt{1-\bar\alpha_t}} \epsilon_\theta(X_t, t) + \sigma_t z\right)
    $
    \State \textcolor{blue}{/* Combine causal and environmental patches */}
    \State 
    $
    X_{t-1} = m \odot X_{t-1}^{cau} +  (1-m) \odot X_{t-1}^{env}
    $
    \vspace{0.5mm}
    \State 
    $
    X_t \sim \mathcal{N}(\sqrt{1 - \beta_{t-1}}X_{t-1}, \beta_{t-1}I)
    $
\EndFor
\State \textbf{return} \( X_A \) as the augmentation dataset
\end{algorithmic}
\end{algorithm}
\end{center}

The algorithm for \textbf{Causal Intervention with Diffusion Inpainting} aims to augment ST observation data through a series of diffusion steps that iteratively refine the data by applying causal interventions and combining them with environmental patches. Here is a detailed step-by-step description:

\begin{itemize}
    \item \textbf{Input:} The original ST observation data $X$, and a masked image $X_{\text{mask}}$.
    \item \textbf{Output:} An augmented ST observation dataset $X_A$.
    \item The process begins by initializing $X_T$, which represents the data at the final diffusion step, to be a sample from a normal distribution centered at zero with identity covariance.
    \item The main loop of the algorithm runs backward from the last diffusion step $T$ to the first. In each step:
    \begin{enumerate}
        \item Gaussian noise $\epsilon_t$ is sampled to simulate the diffusion process.
        \item A causal region $X_{\text{cau}}$ is sampled where the causal effect is calculated as a blend of the original data and the Gaussian noise, emphasizing areas of interest that should retain more original data characteristics.
        \item Gaussian noise $N_t$ is sampled again, providing variability to the non-causal or environmental regions.
        \item The environmental patches $X_{\text{env}}$ are updated using the data from the previous step adjusted by a damping factor and the added noise, simulating environmental changes.
        \item The causal and environmental patches are then combined, where the mask $M$ determines the specific locations for the causal and environmental updates in the data, specifying which parts are from the causal region and which are from the environmental region.
        \item The data for the next step, $X_{t-1}$, is computed by normalizing the combined updates, preparing it for the next iteration or output if it is the first step.
    \end{enumerate}
    \item Finally, the algorithm outputs the augmented data set $X_A$, which is the result of the iterative causal intervention and environmental blending over the diffusion process.
\end{itemize}

 \section{Details of experiments}\label{app:b}
\textbf{SSIM} stands for Structural Similarity Index Measure, which is a method for measuring the similarity between two images. It compares the structural information of the images, including luminance, contrast, and texture, to determine how similar they are. SSIM is commonly used in image and video processing applications, such as image compression and quality assessment.

\textbf{PSNR} stands for Peak Signal-to-Noise Ratio. It is a measure of video or image quality that compares the original signal to the compressed or transmitted signal. The higher the PSNR value, the better the quality of the compressed or transmitted signal. PSNR is commonly used in video and image compression applications to evaluate the effectiveness of compression algorithms.
    
\textbf{MSE (Mean Squared Error)} loss is a commonly used loss function in machine learning and deep learning models. This loss function calculates the average of the squared differences between the predicted and actual values.  


\textbf{Datasets.} Here we summarize the details (Tab. \ref{tab:main}) of the datasets used in this paper:
\begin{itemize}[leftmargin=*]

    \item TaxiBJ+: This dataset contains trajectory data obtained from the GPS of taxis in Beijing, divided into two separate channels: inflow and outflow. Additionally, the dataset has been extended from 32×32 to 128×128 by collecting recent trajectory data from Beijing.

    \item KTH: This dataset includes 25 individuals performing six different actions: walking, jogging, running, boxing, waving, and clapping. The complexity of human movements arises from the unique variations each individual displays while executing these actions. By examining previous frames, the model can understand the subtleties of human dynamics and predict future extended postural changes.

    \item SEVIR: This dataset consists of weather images that have been sampled and aligned using radar and satellite data. It is designed as a foundational resource to support algorithm development in meteorological research.
    
    \item DRS: This dataset describes the diffusion process of nonlinear wave, which satisfies the diffusion equation. 

    \item FireSys: The FireSys dataset comprises data associated with fire observations, capturing both temporal and spatial trends of fire evolution, which faithfully represent the progression status in a natural setting.
     
\end{itemize}

\section{Broader Impact}\label{app:c}

The development and application of the CaPaint framework in spatio-temporal (ST) dynamics bring several positive broader impacts. Understanding these impacts is crucial for responsible AI research and deployment.

\textbf{1. Data Imputation in Sparse Scenarios:}
CaPaint excels in sparse data scenarios, effectively filling in missing data. This reduces the need for extensive sensor deployments, significantly lowering the cost associated with sensor installation. By optimizing data coverage and utilization, CaPaint not only enhances resource efficiency but also achieves substantial cost savings.

\textbf{2. Enhanced Predictive Accuracy and Interpretability:}
CaPaint can identify and intervene in non-causal regions, improving the predictive accuracy and interpretability in various ST domains such as meteorology, human mobility, and disaster management. This improvement leads to better decision-making processes and resource allocation, ultimately benefiting society by providing more reliable and understandable predictive models.

\textbf{3. Cost-Effective Solutions:}
By reducing the complexity of optimal ST causal discovery models, CaPaint offers a cost-effective solution for handling high-dimensional ST data. This makes advanced predictive technologies more accessible across a broader range of applications, particularly in fields with limited computational resources.

\textbf{4. Promotion of Causal Reasoning in AI:}
The integration of causal reasoning into ST models encourages the development of AI systems that better mimic human understanding of cause-and-effect relationships. This can lead to more robust AI models capable of generalizing across different scenarios, fostering trust and reliability in AI applications.

\textbf{5. Innovation in Data Augmentation Techniques:}
CaPaint introduces novel data augmentation methods using diffusion inpainting, which can inspire further research and innovation in data augmentation and ST prediction. This can lead to the emergence of new techniques, enhancing the robustness and performance of AI models in various domains.

The CaPaint framework represents a significant advancement in the field of ST dynamics, particularly in its ability to address sparse data scenarios, which reduces the need for extensive sensor deployments and lowers associated costs. Additionally, CaPaint enhances predictive accuracy, interpretability, and efficiency, promotes causal reasoning in AI, and introduces innovative data augmentation techniques. Responsible AI research and deployment should leverage these strengths to maximize benefits while minimizing risks.

\section{Metrics}\label{metrics}

In our research, we investigate the performance of our models using Mean Squared Error (MSE), Mean Absolute Error (MAE), and Structural Similarity Index Measure (SSIM). The formulas for evaluating these indicators, converted into decibels (dB) where applicable, are as follows:

\subsection*{Mean Squared Error (MSE)}
Mean Squared Error (MSE) measures the average of the squares of the errors, that is, the average squared difference between the estimated values and the actual value. The MSE is given by:
\begin{equation}
\text{MSE} = \frac{1}{N} \sum_{i=1}^{N} (Y_i - \hat{Y_i})^2
\end{equation}
where $Y_i$ is the actual value, $\hat{Y_i}$ is the predicted value, and $N$ is the number of observations.

\subsection*{Mean Absolute Error (MAE)}
Mean Absolute Error (MAE) measures the average magnitude of the errors in a set of predictions, without considering their direction. It is the average over the test sample of the absolute differences between prediction and actual observation where all individual differences have equal weight. The MAE is given by:
\begin{equation}
\text{MAE} = \frac{1}{N} \sum_{i=1}^{N} \left| Y_i - \hat{Y_i} \right|
\end{equation}
where $Y_i$ is the actual value, $\hat{Y_i}$ is the predicted value, and $N$ is the number of observations.

\subsection*{Structural Similarity Index Measure (SSIM)}
Structural Similarity Index Measure (SSIM) is used for measuring the similarity between two images. The SSIM index is a decimal value between -1 and 1, where 1 is only reachable in the case of two identical sets of data. The SSIM formula can be quite complex due to its consideration of luminance, contrast, and structure comparison functions between the two images:
\begin{equation}
\text{SSIM}(x, y) = \frac{(2\mu_x \mu_y + C_1)(2\sigma_{xy} + C_2)}{(\mu_x^2 + \mu_y^2 + C_1)(\sigma_x^2 + \sigma_y^2 + C_2)}
\end{equation}
where $\mu_x$, $\mu_y$ are the average of $x$ and $y$ respectively, $\sigma_x^2$, $\sigma_y^2$ are the variance of $x$ and $y$ respectively, $\sigma_{xy}$ is the covariance of $x$ and $y$, and $C_1, C_2$ are variables to stabilize the division with weak denominator.

\section{Limitations}\label{limitation}

While the implementation of the CaPaint method has demonstrated significant improvements in prediction accuracy and detail preservation in spatio-temporal forecasting tasks, its enhancements are most pronounced in scenarios characterized by data scarcity or uneven data distribution. In contexts where datasets are abundant and exhibit a broad and uniform distribution, the incremental gains offered by CaPaint may not be as substantial. Nevertheless, the method remains effective, providing consistent, albeit smaller, improvements across diverse data environments. 

\section{An example of ST Inpainting on SEVIR}\label{app:e}

\begin{figure}[h!]
    \centering
    \includegraphics[width=1\textwidth]{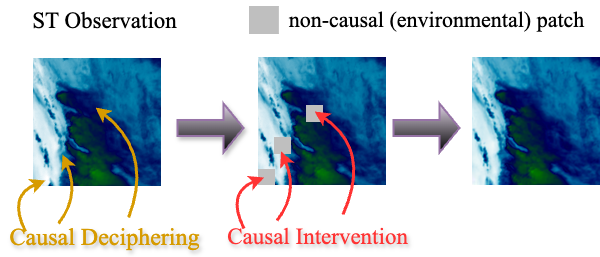}
    \caption{Inpainting Example of our proposed CaPaint.}
\end{figure}

The figure illustrates the process of maintaining causal regions intact while performing inpainting on non-causal (environmental) regions. The approach involves identifying and deciphering the causal regions (left), intervening by applying diffusion inpainting on the environmental patches (middle), and subsequently generating altered ST data copies (right). This method ensures that the intrinsic causal relationships within the data are preserved, while variations are introduced in the environmental context to augment the dataset effectively.

\section{Uneven Distribution of Sensors Leading to Data Scarcity in Global Oceanic Observation}

\begin{figure}[h]
  \centering
  \includegraphics[width=0.95\linewidth]{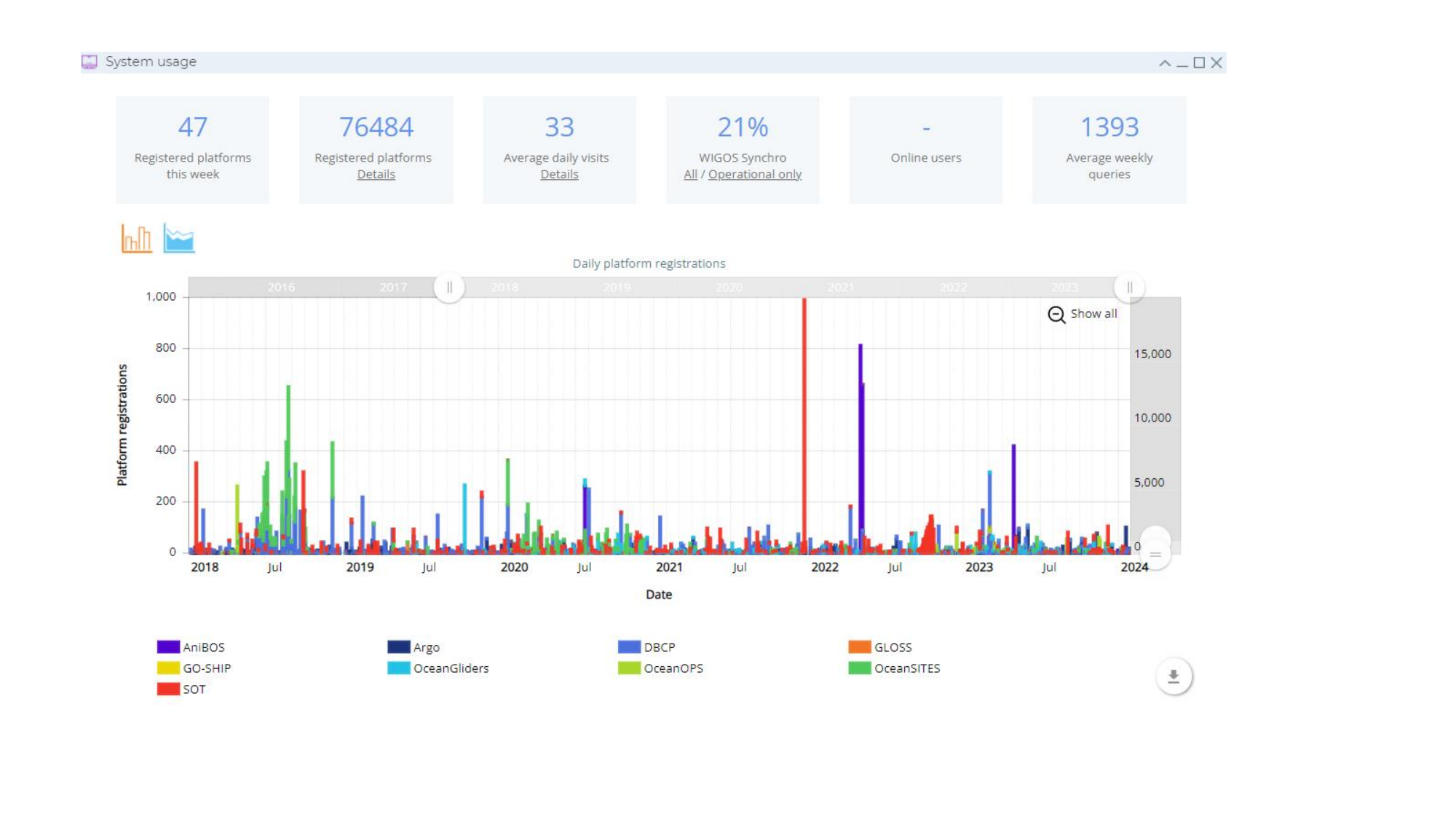}\vspace{-4pt}
  \caption{Temporal distributional heterogeneity within the global oceanic observation platforms, which reveals that there are pronounced disparities in the deployment numbers of various types of sensors during different time intervals.}
  \label{fig:exp2}
  \vspace{-7pt}
\end{figure}

\section{Experimental Parameters}\label{app:g}

In this experiment, we employ different deep learning models and optimize them for training. All experiments are conducted on hardware equipped with 24 NVIDIA GeForce RTX 4090 GPUs. The optimizer used is Adam, and different learning rates (LR) and batch sizes are set for each model. The specific parameter settings are shown in the table below:

\begin{table}[h!]
\centering
\begin{tabular}{|c|c|c|}
\hline
\textbf{Model} & \textbf{Learning Rate (LR)} & \textbf{Batch Size} \\
\hline
CLSTM & 0.001 & 8 \\
\hline
MAU & 0.001 & 8 \\
\hline
MMVP & 0.004 & 4 \\
\hline
PredRNNv2 & 0.001 & 8 \\
\hline
SimVP & 0.004 & 4 \\
\hline
ViT & 0.004 & 4 \\
\hline
Earthfarsser & 0.001 & 8 \\
\hline
\end{tabular}
\caption{Learning rates and batch sizes for different backbones}
\label{tab:experiment_params}
\end{table}

These parameter settings are chosen based on the characteristics of each model and preliminary experimental results on the validation set, aiming to optimize the training efficiency and performance of the models. The Adam optimizer is used with a OneCycle learning rate scheduler, where the maximum learning rate is set according to the specified learning rate for each model, and the number of steps per epoch and the total number of epochs are set based on the training data and experimental setup. During the experiments, we ensure that all models are trained under the same hardware conditions to guarantee the comparability and reproducibility of the results.

\section{Visualizations on KTH}
\begin{figure}[h]
  \centering
  \includegraphics[width=0.95\linewidth]{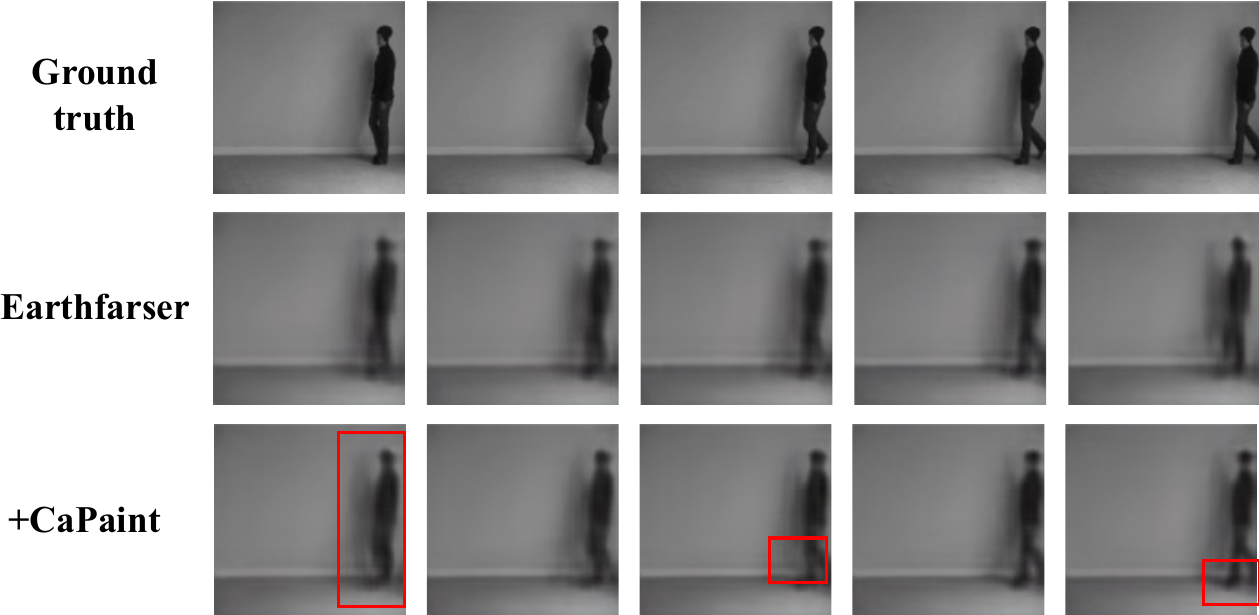}\vspace{-4pt}
  \caption{Visualizations on KTH dataset showing the last 5 frames}
  \vspace{-7pt}
\end{figure}

The first row shows the ground truth for a walking individual. The second row, processed by Earthfarser, exhibits noticeable blurring and loss of detail. The third row, enhanced with +CaPaint, demonstrates a marked improvement in capturing fine details such as the shadow of the person and the accuracy of the foot motion, as highlighted in the red boxes.

\section{Visualizations on Diffusion Reaction System}

\begin{figure}[h]
  \centering
  \includegraphics[width=0.95\linewidth]{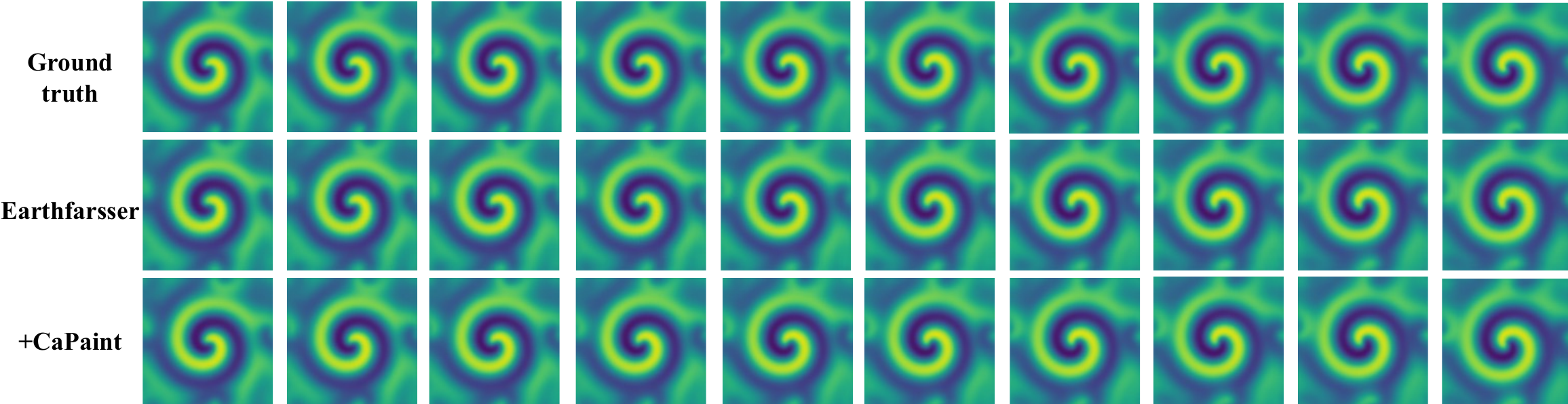}\vspace{-4pt}
  \caption{Visualizations on DRS dataset showing 10 frames}
  \vspace{-7pt}
\end{figure}

The introduction of CaPaint has led to reductions in Mean Squared Error MSE and MAE, while the SSIM has shown improvements. These changes indicate that the CaPaint method effectively enhances model prediction accuracy and image quality. However, due to the high quality of the model predictions, the improvements might not be readily observable to the naked eye. Despite this, the positive effects of CaPaint are clearly evident through quantitative metrics, demonstrating its potential and practicality in enhancing the accuracy of complex dynamic systems predictions.

\newpage

\section*{NeurIPS Paper Checklist}

\begin{enumerate}

\item {\bf Claims}
    \item[] Question: Do the main claims made in the abstract and introduction accurately reflect the paper's contributions and scope?
    \item[] Answer: \answerYes{} 
    \item[] Justification: In this paper, we introduce the spatio-temporal causal concept in the data mining realm, aimed at enhancing the reliability and accuracy of financial spatio-temporal prediction.

\item {\bf Limitations}
    \item[] Question: Does the paper discuss the limitations of the work performed by the authors?
    \item[] Answer: \answerYes{} 
    \item[] Justification: In this work, we systematically discuss the limitations of our research and outline directions for future work.

\item {\bf Theory Assumptions and Proofs}
    \item[] Question: For each theoretical result, does the paper provide the full set of assumptions and a complete (and correct) proof?
    \item[] Answer: \answerNA{} 
    \item[] Justification: This paper does not include experimental results related to theoretical aspects.

    \item {\bf Experimental Result Reproducibility}
    \item[] Question: Does the paper fully disclose all the information needed to reproduce the main experimental results of the paper to the extent that it affects the main claims and/or conclusions of the paper (regardless of whether the code and data are provided or not)?
    \item[] Answer: \answerYes{} 
    \item[] Justification: We provide the code necessary for replicating the studies described in this paper via an anonymous link, and we detail the experimental setup for the replication in the article itself.

\item {\bf Open access to data and code}
    \item[] Question: Does the paper provide open access to the data and code, with sufficient instructions to faithfully reproduce the main experimental results, as described in supplemental material?
    \item[] Answer: \answerYes{} 
    \item[] Justification: For the datasets disclosed in the article, we have provided information regarding their sources and origins.

\item {\bf Experimental Setting/Details}
    \item[] Question: Does the paper specify all the training and test details (e.g., data splits, hyperparameters, how they were chosen, type of optimizer, etc.) necessary to understand the results?
    \item[] Answer: \answerYes{} 
    \item[] Justification: we have specified all the training and test details (e.g., data splits, hyperparameters, how they were chosen, type of optimizer, etc.) necessary to understand the results.

\item {\bf Experiment Statistical Significance}
    \item[] Question: Does the paper report error bars suitably and correctly defined or other appropriate information about the statistical significance of the experiments?
    \item[] Answer: \answerYes{} 
    \item[] Justification: In this paper, we have reported error bars suitably and correctly defined or other appropriate information about the statistical significance of the experiments.

\item {\bf Experiments Compute Resources}
    \item[] Question: For each experiment, does the paper provide sufficient information on the computer resources (type of compute workers, memory, time of execution) needed to reproduce the experiments?
    \item[] Answer: \answerYes{} 
    \item[] Justification: In this paper, we provide detailed information about the experimental resources, including GPU configurations used in our studies.
    
\item {\bf Code Of Ethics}
    \item[] Question: Does the research conducted in the paper conform, in every respect, with the NeurIPS Code of Ethics \url{https://neurips.cc/public/EthicsGuidelines}?
    \item[] Answer: \answerYes{} 
    \item[] Justification: The study presented in this paper conforms to the NeurIPS Code of Ethics.

\item {\bf Broader Impacts}
    \item[] Question: Does the paper discuss both potential positive societal impacts and negative societal impacts of the work performed?
    \item[] Answer: \answerYes{} 
    \item[] Justification: We have provided the societal impacts of the work.
    
\item {\bf Safeguards}
    \item[] Question: Does the paper describe safeguards that have been put in place for responsible release of data or models that have a high risk for misuse (e.g., pretrained language models, image generators, or scraped datasets)?
    \item[] Answer: \answerNA{} 
    \item[] Justification: This paper does not address issues related to this aspect.

\item {\bf Licenses for existing assets}
    \item[] Question: Are the creators or original owners of assets (e.g., code, data, models), used in the paper, properly credited and are the license and terms of use explicitly mentioned and properly respected?
    \item[] Answer: \answerYes{} 
    \item[] Justification: All creators and original owners of the assets used in our paper, such as code, data, and models, have been properly credited. We have explicitly mentioned the licenses and terms of use for each asset and have ensured full compliance with these terms throughout our research.

\item {\bf New Assets}
    \item[] Question: Are new assets introduced in the paper well documented and is the documentation provided alongside the assets?
    \item[] Answer: \answerNA{} 
    \item[] Justification: The research presented in this paper is not concerned with new assets.

\item {\bf Crowdsourcing and Research with Human Subjects}
    \item[] Question: For crowdsourcing experiments and research with human subjects, does the paper include the full text of instructions given to participants and screenshots, if applicable, as well as details about compensation (if any)? 
    \item[] Answer: \answerNA{} 
    \item[] Justification: This paper does not involve experiments or research related to human subjects.

\item {\bf Institutional Review Board (IRB) Approvals or Equivalent for Research with Human Subjects}
    \item[] Question: Does the paper describe potential risks incurred by study participants, whether such risks were disclosed to the subjects, and whether Institutional Review Board (IRB) approvals (or an equivalent approval/review based on the requirements of your country or institution) were obtained?
    \item[] Answer: \answerNA{} 
    \item[] Justification: This paper does not address potential risks incurred by study participants.

\end{enumerate}

\end{document}